\documentclass[lettersize,journal]{IEEEtran}
\usepackage{amsmath,amsfonts}
\usepackage{algorithmic}
\usepackage{algorithm}
\usepackage{array}
\usepackage[caption=false,font=normalsize,labelfont=sf,textfont=sf]{subfig}
\usepackage{textcomp}
\usepackage{stfloats}
\usepackage{url}
\usepackage{verbatim}
\usepackage{graphicx}
\usepackage{cite}
\usepackage[accsupp]{axessibility} 
\usepackage[bottom]{footmisc}
\usepackage{booktabs}
\usepackage{multirow}
\usepackage{graphicx}
\usepackage{xcolor}
\newcommand{\mysmall}[1]{\scriptsize{\color{gray}{#1}}}

\usepackage[normalem]{ulem}
\useunder{\uline}{\ul}{}

\begin{document}

\title{SyncTalk++: High-Fidelity and Efficient Synchronized Talking Heads Synthesis \\ Using Gaussian Splatting}

\author{Ziqiao Peng, Wentao Hu, Junyuan Ma, Xiangyu Zhu,~\IEEEmembership{Senior Member,~IEEE,} Xiaomei Zhang, Hao Zhao,~\IEEEmembership{Member,~IEEE,} Hui Tian,~\IEEEmembership{Senior Member,~IEEE,} Jun He,~\IEEEmembership{Member,~IEEE,} Hongyan Liu\textsuperscript{*},~\IEEEmembership{Member,~IEEE,} Zhaoxin Fan\textsuperscript{*}

\thanks{Hongyan Liu and Zhaoxin Fan are the corresponding authors.}
\thanks{Ziqiao Peng and Jun He are with the School of Information, Renmin University of China.}
\thanks{Wentao Hu and Hui Tian are with the School of Information and Communication Engineering, Beijing University of Posts and Telecommunications.}
\thanks{Junyuan Ma is with the Aerospace Information Research Institute, Chinese Academy of Sciences.}
\thanks{Xiangyu Zhu, and Xiaomei Zhang are with the Institute of Automation, Chinese Academy of Sciences.}
\thanks{Hao Zhao is with the Institute for AI Industry Research, Tsinghua University.}
\thanks{Hongyan Liu is with the School of Economics and Management, Tsinghua University.}
\thanks{Zhaoxin Fan is with the Beijing Advanced Innovation Center for Future Blockchain and Privacy Computing, School of Artificial Intelligence, Beihang University, Hangzhou International Innovation Institute, Beihang University.}

}

\markboth{Journal of \LaTeX\ Class Files,~Vol.~14, No.~8, August~2021}%
{Shell \MakeLowercase{\textit{et al.}}: A Sample Article Using IEEEtran.cls for IEEE Journals}

\maketitle

\begin{abstract}

Achieving high synchronization in the synthesis of realistic, speech-driven talking head videos presents a significant challenge. 
A lifelike talking head requires synchronized coordination of subject identity, lip movements, facial expressions, and head poses. The absence of these synchronizations is a fundamental flaw, leading to unrealistic results.
To address the critical issue of synchronization, identified as the ``devil'' in creating realistic talking heads, we introduce SyncTalk++, which features a Dynamic Portrait Renderer with Gaussian Splatting to ensure consistent subject identity preservation and a Face-Sync Controller that aligns lip movements with speech while innovatively using a 3D facial blendshape model to reconstruct accurate facial expressions. To ensure natural head movements, we propose a Head-Sync Stabilizer, which optimizes head poses for greater stability. Additionally, SyncTalk++ enhances robustness to out-of-distribution (OOD) audio by incorporating an Expression Generator and a Torso Restorer, which generate speech-matched facial expressions and seamless torso regions. Our approach maintains consistency and continuity in visual details across frames and significantly improves rendering speed and quality, achieving up to 101 frames per second. Extensive experiments and user studies demonstrate that SyncTalk++ outperforms state-of-the-art methods in synchronization and realism. We recommend watching the supplementary video: https://ziqiaopeng.github.io/synctalk++.

\end{abstract}


\section{Introduction}
\label{sec:intro}
The need to generate dynamic and realistic speech-driven talking heads has intensified, driven by emerging applications in domains such as digital assistants~\cite{thies2020neural,yang2025megadance}, virtual reality~\cite{peng2023selftalk,zhou2024meta}, and filmmaking~\cite{kim2018deep,peng2025dualtalk,wu2024vgg}. These applications demand high visual fidelity and seamless integration of critical synchronized factors, including subject identity, lip movements, facial expressions, and head poses. The ultimate goal is to create synthetic videos that are indistinguishable from real human captures, thereby aligning with human perceptual expectations and enabling more expressive communication.

At the core of realistic talking head synthesis lies the challenge of synchronization across critical factors. These components must be perfectly aligned to produce a coherent and lifelike representation. However, the inherent ambiguity in mapping speech to facial movements introduces significant challenges, often resulting in artifacts that disrupt the perceived realism. This ambiguity makes it difficult to achieve an accurate and consistent depiction of facial dynamics based solely on speech. Synchronization in talking head synthesis is particularly critical because of the way humans process and interpret facial movement in communication. Facial expressions and lip movements are tightly coupled with speech, and any misalignment between these factors can disrupt the perception of realism. Thus, addressing the synchronization challenge involves dissecting the ambiguity in audio-visual mappings, turning this “devil” in the details into a focal point for ensuring good fidelity in talking head synthesis.

Current methods for generating talking heads are generally divided into two main categories: 2D generation and 3D reconstruction methods. 2D generation methods, including Generative Adversarial Networks (GAN)\cite{goodfellow2020generative,zhang2023dinet,wang2023seeing,guan2023stylesync,zhong2023identity,tan2023emmn, guo2024liveportrait, tan2024flowvqtalker,wu2023ganhead } and recent diffusion models\cite{tian2024emo,ma2023dreamtalk,shen2023difftalk}, have shown significant progress in modeling lip movements and generating talking heads from single images. These methods, trained on large datasets, excel at producing realistic head movements and facial expressions. However, their reliance on 2D information limits their ability to achieve accurate synchronization across critical factors. Without three-dimensional prior knowledge, these methods often produce facial movements that do not conform to physical laws, leading to inconsistencies such as variations in facial features across frames. These inconsistencies arise from the 2D models’ inability to capture the depth and spatial relationships necessary for realistic facial animation, resulting in outputs that may lack identity consistency and exhibit artifacts. 

Similarly, the emerging 3D reconstruction methods in recent years, such as those based on Neural Radiance Fields (NeRF) \cite{mildenhall2021nerf,guo2021ad,yao2022dfa,shen2022learning,ye2023geneface,li2023efficient, zhang2024learning,chu2024gpavatar, li2024er} and Gaussian Splatting \cite{kerbl20233d,yu2024gaussiantalker,li2024talkinggaussian}, have shown excellent performance in maintaining identity consistency between frames and preserving facial details. These methods utilize ray and point information in three-dimensional space to generate high-fidelity head models, ensuring continuity and realism of the character from different perspectives.
However, these 3D reconstruction methods also face some challenges. They struggle to achieve highly synchronized lip movements with only a limited volume of 4-5 minutes of training data. Most existing methods use pre-trained models like DeepSpeech~\cite{amodei2016deep} for automatic speech recognition to extract audio features. However, the feature distribution from speech-to-text differs from the speech-to-image distribution needed for this task, often resulting in lip movements that do not match the speech.

\begin{figure*}
\begin{center}
   \includegraphics[width=1.\linewidth]{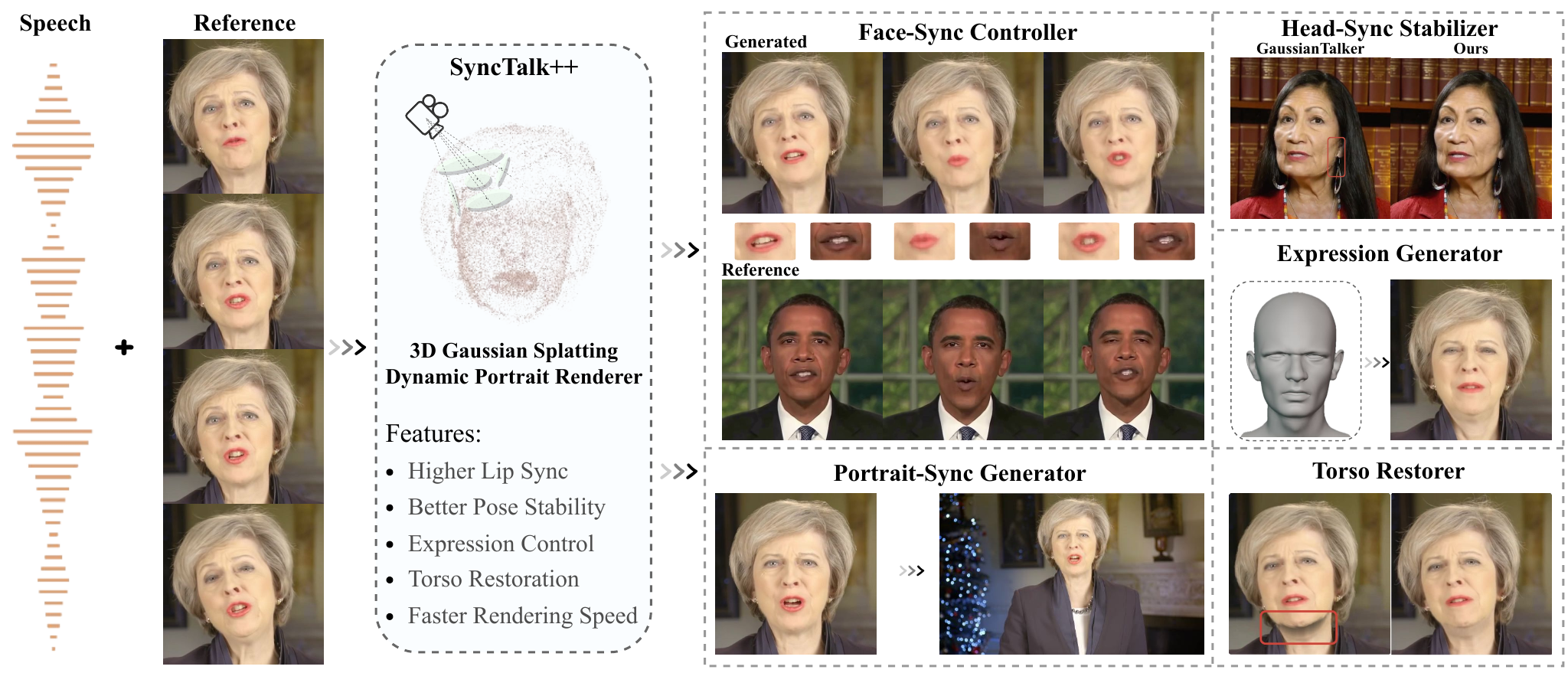}
\end{center}
   \caption{The proposed SyncTalk++ uses 3D Gaussian Splatting for rendering. It can generate synchronized lip movements, facial expressions, and more stable head poses, and features faster rendering speeds while applying to high-resolution talking videos.}
\label{fig:1}
\vspace{-1em}
\end{figure*}

Based on the above motivations, we find that the ``devil'' is in the synchronization. Existing methods need more synchronization in four key areas: subject identity, lip movements, facial expressions, and head poses. To address these synchronization challenges, we propose three key sync modules: the Face-Sync Controller, the Head-Sync Stabilizer, and the Dynamic Portrait Renderer, as shown in Fig.~\ref{fig:1}.

The first is the synchronization of lip movements and facial expressions, we use the Face-Sync Controller, which employs an audio-visual encoder and a 3D facial blendshape model to achieve high synchronization. Unlike traditional methods that rely on pre-trained ASR models, our approach leverages an audio-visual synchronization encoder trained specifically for aligning audio features with lip movements. This ensures that the extracted audio features better aligned with the movements of the lips. The Face-Sync Controller also incorporates a 3D facial blendshape model, which utilizes semantically meaningful facial coefficients to capture and control expressions. This allows the system to produce more nuanced and realistic facial expressions independent of lip movements. 

The second is the synchronization of head poses, the Head-Sync Stabilizer plays a vital role in maintaining stability. This module employs a two-stage optimization framework that starts with an initial estimation of head pose parameters and is followed by a refined process integrating optical flow information and keypoint tracking. By using a semantic weighting module to reduce the weight of unstable points such as eyebrow and eye movements, the Head-Sync Stabilizer significantly improves the accuracy and stability of head poses, ensuring that head movements remain natural and consistent. 

The third is the synchronization of subject identity, the Dynamic Portrait Renderer takes charge of high-fidelity facial rendering and the restoration of fine details. Utilizing 3D Gaussian Splatting, the renderer explicitly models 3D Gaussian primitives, allowing for high-fidelity reconstruction of facial features from multiple perspectives. This method not only improves rendering speed but also reduces visual artifacts. 

In real-world applications of talking heads, commonly used out-of-distribution (OOD) audio, such as audio from different speakers or text-to-speech (TTS) systems, often leads to mismatches between facial expressions and spoken content. For instance, an OOD audio might make a character frown during a cheerful topic, reducing the realism of the video. To address this issue, we introduce the OOD Audio Expression Generator. This module creates facial expressions that match the speech content, which we call speech-matched expressions, enhancing the realism of the expressions even with OOD audio. Additionally, to handle the limited generalization of Gaussian Splatting with unseen data, we incorporate a codebook that minimizes cross-modal mapping uncertainties. Additionally, when generating videos with OOD audio, inconsistencies may arise, such as the character’s mouth being open in the original frame but closed in the generated frame. This discrepancy in jaw position can lead to pixel gaps between the generated head and torso. To address this, we introduce the Torso Restorer, which uses a lightweight U-Net-based inpainting model. This module effectively bridges these gaps, ensuring seamless integration of the head and torso, thus improving the final video's overall visual quality and coherence.

The contributions of this paper are summarized below:
\begin{itemize}
\item[$\bullet$] We present SyncTalk++, a talking head synthesis method using Gaussian Splatting, achieving high synchronization of identity, lip movements, expressions, and head poses, with 101 frames per second rendering and improved visual quality.

\item[$\bullet$] We enhance the robustness for out-of-distribution (OOD) audio inference by using an Expression Generator and Torso Restorer to generate speech-matched facial expressions and repair artifacts at head-torso junctions.

\item[$\bullet$] We compare our method with recent state-of-the-art methods, and both qualitative and quantitative comparisons demonstrate that our method outperforms existing methods and is ready for practical deployment.
\end{itemize}

A preliminary version of this work was presented in~\cite{peng2024synctalk}. In this extended work, we make improvements in four aspects: (1) We adopt Gaussian Splatting to replace NeRF implicit modeling, achieving faster rendering speed and higher fidelity; (2) We introduce an Expression Generator and Torso Restorer to enhance robustness against out-of-distribution (OOD) audio, thereby improving stability in practical applications; (3) We optimize the facial tracking module by incorporating a semantic weighting module to improve reconstruction stability; and (4) We conduct broader and more comprehensive experiments, demonstrating that our method outperforms the existing state-of-the-art.

\section{Related Work}
\label{sec:related}

\subsection{2D Generation-based Talking Head Synthesis}

\subsubsection{GAN-based Talking Head Synthesis}

Recently, GAN-based talking head synthesis~\cite{chen2018lip,kr2019towards,chen2019hierarchical,zhou2019talking,das2020speech,vougioukas2020realistic,meshry2021learned,zhou2021pose,song2022everybody} has emerged as an essential research area in computer vision. For example, Wav2Lip~\cite{prajwal2020lip} introduces a lip synchronization expert to supervise lip movements, enforcing the consistency of lip movements with the audio. IP-LAP~\cite{zhong2023identity} proposes a two-stage framework consisting of audio-to-landmark generation and landmark-to-video rendering procedures, surpassing wav2lip and similar methods in video generation quality and alleviating the poor fusion of generated lip region images with facial images. These methods generate only the lower half of the face or the lip region, while the other areas remain the original video content. This can lead to uncoordinated facial movements, and artifacts are likely to appear at the edge between the original and generated regions. Methods like~\cite{chen2019hierarchical, zhou2020makelttalk, lu2021live, wang2021audio2head}  generate the entire face but struggle to maintain the original facial details. Apart from video stream techniques, efforts have also been made to enable a single image to ``speak" using speech. For example, SadTalker~\cite{zhang2023sadtalker} uses 3D motion coefficients derived from audio to modulate a 3D-aware face render implicitly. 

\subsubsection{Diffusion-based Talking Head Synthesis}
With the widespread application of diffusion models in the Artificial Intelligence Generated Content field, their excellent generative capabilities have also been utilized for talking head synthesis, such as~\cite{tian2024emo, shen2023difftalk, xu2024hallo, chen2024echomimic,peng2025omnisync}. For example, EMO~\cite{tian2024emo} employs Stable Diffusion~\cite{rombach2022high} as the foundational framework to achieve vivid video synthesis of a single image.
DiffTalk~\cite{shen2023difftalk}, in addition to using audio conditions to drive the lip motions, further incorporates reference images and facial landmarks as extra driving factors for personalized facial modeling. Hallo~\cite{xu2024hallo} introduces a hierarchical cross-attention mechanism to augment the correlation between audio inputs and non-identity-related motions. However, these methods rely solely on a single reference image to synthesize a series of continuous frames, making it difficult to maintain a single character's identity consistently. This often results in inconsistencies in teeth and lips. The lack of 3D facial structure information can sometimes lead to distorted facial features. Additionally, these diffusion model-based methods often require significant computational resources, which presents challenges in deployment. 

Compared to these methods, SyncTalk++ uses Gaussian Splatting to perform three-dimensional modeling of the face. Its capability to represent continuous 3D scenes in canonical spaces translates to exceptional performance in maintaining subject identity consistency and detail preservation. Simultaneously, its training and rendering speed is significantly superior to 2D-based methods. 

\begin{figure*}
\begin{center}
   \includegraphics[width=1.\linewidth]{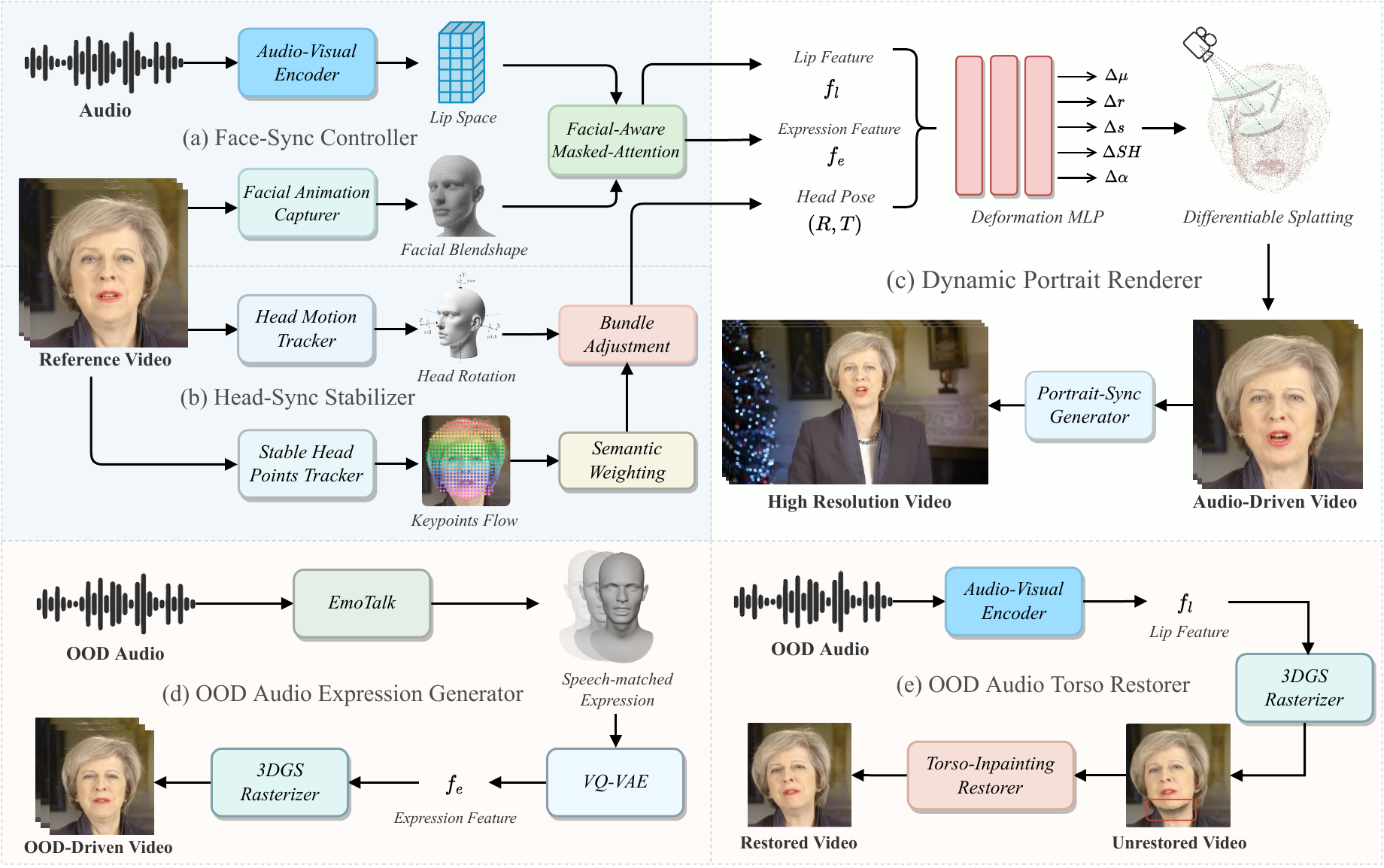}
\end{center}
   \caption{\textbf{Overview of SyncTalk++.} Given a cropped reference video of a talking head and the corresponding speech, SyncTalk++ can extract the Lip Feature $f_l$, Expression Feature $f_e$, and Head Pose $(R, T)$ through two synchronization modules $(a)$ and $(b)$. Then, Gaussian Splatting is used to model and deform the head, producing a talking head video. The OOD Audio Expression Generator and Torso Restorer can generate speech-matched facial expressions and repair artifacts at head-torso junctions.}
\label{fig:2}
\end{figure*}

\subsection{3D Reconstruction-based Talking Head Synthesis}
\subsubsection{NeRF-based Talking Head Synthesis}

With the recent rise of NeRF, numerous fields have begun to utilize it to tackle related challenges~\cite{martin2021nerf,gao2022nerf}.
Previous work~\cite{guo2021ad,yao2022dfa,liu2022semantic,ye2023geneface} has integrated NeRF into the task of synthesizing talking heads and have used audio as the driving signal, but these methods are all based on the vanilla NeRF model. For instance, AD-NeRF~\cite{guo2021ad} requires approximately 10 seconds to render a single image. RAD-NeRF~\cite{tang2022real} aims for real-time video generation and employs a NeRF based on Instant-NGP~\cite{muller2022instant}. ER-NeRF~\cite{li2023efficient} innovatively introduces triple-plane hash encoders to trim the empty spatial regions, advocating for a compact and accelerated rendering approach. GeneFace~\cite{ye2023geneface} attempts to reduce NeRF artifacts by translating speech features into facial landmarks, but this often results in inaccurate lip movements. Portrait4D~\cite{deng2024portrait4d} creates pseudo-multi-view videos from existing monocular videos and trains on a large-scale multi-view dataset. It can reconstruct multi-pose talking heads from a single image. However, it cannot be directly driven by speech and faces the same problem as 2D one-shot methods in maintaining identity consistency. Attempts to create character avatars with NeRF-based methods, such as \cite{gafni2021dynamic,zheng2022avatar,zielonka2023instant,zheng2023pointavatar}, cannot be directly driven by speech.
These methods only use audio as a condition, without a clear concept of sync, and usually result in average lip movement. Additionally, previous methods lack control over facial expressions, being limited to controlling blinking only, and cannot model actions like raising eyebrows or frowning.

\subsubsection{3DGS-based Talking Head Synthesis}

Recently, Gaussian Splatting based on explicit parameter modeling has demonstrated excellent performance in 3D rendering~\cite{kerbl20233d,lu2024scaffold,yu2024mip}. 3DGS has been explored for application in 3D human avatar modeling. 3DGS-Avatar~\cite{qian20243dgs} utilizes 3D Gaussian projection and a non-rigid deformation network to quickly generate animatable human head avatars from monocular videos. GauHuman~\cite{hu2024gauhuman} combines the LBS weight field module and the posture refinement module to transform 3D Gaussian distribution from the canonical space to the posture space. PSAvatar~\cite{zhao2024psavatar} uses point-based deformable shape model (PMSM) and 3D Gaussian modeling to excel in real-time animation through flexible and detailed 3D geometric modeling. GaussianAvatars~\cite{qian2024gaussianavatars} innovatively combines the FLAME mesh model with 3D Gaussian distribution to achieve detailed head reconstruction through the spatial properties of the Gaussian distribution. Gaussian head avatar~\cite{xu2024gaussian} utilizes controllable 3D Gaussian models for high-fidelity head avatar modeling. GaussianTalker~\cite{cho2024gaussiantalker} integrates 3D Gaussian attributes with audio features into a shared implicit feature space, using 3D Gaussian splatting for fast rendering. It is a real-time pose-controllable talking head model that significantly improves facial realism, lip synchronization accuracy, and rendering speed. TalkingGaussian~\cite{li2024talkinggaussian} is a deformation-based framework leveraging the point-based Gaussian Splatting to represent facial movements by maintaining a stable head structure and smoothly, continuously deforming Gaussian primitives, thereby generating high-fidelity talking head avatars. However, these methods have certain limitations in synchronization mechanisms, such as the inability to consistently maintain stable head poses, leading to the separation of the head and torso.

In comparison, we use the Face-Sync Controller to capture the relationship between audio and lip movements, thereby enhancing the synchronization of lip movements and expressions, and the Head-Sync Stabilizer to improve head posture stability.

\section{Method}
\label{sec:method}

\subsection{Overview}
In this section, we introduce the proposed SyncTalk++, as shown in Fig.~\ref{fig:2}. SyncTalk++ mainly consists of five parts: a) lip movements and facial expressions controlled by the Face-Sync Controller, b) stable head pose provided by the Head-Sync Stabilizer, c) high-synchronization facial frames rendered by the Dynamic Portrait Renderer, d) speech-matched expressions generated by the OOD Audio Expression Generator, and e) facial and torso fusion details repaired by the OOD Audio Torso Restorer. We will describe the content of these five parts in detail in the following subsections.

\subsection{Face-Sync Controller}
\noindent\textbf{Audio-Visual Encoder.} Existing 3D reconstruction-based methods utilize pre-trained models such as DeepSpeech~\cite{amodei2016deep}, Wav2Vec 2.0~\cite{baevski2020wav2vec}, or HuBERT~\cite{hsu2021hubert}. These are audio feature extraction methods designed for speech recognition tasks. Using an audio encoder designed for Automatic Speech Recognition (ASR) tasks does not truly reflect lip movements. This is because the pre-trained model is based on the distribution of features from audio to text, whereas we need the feature distribution from audio to lip movements.

Considering the above, we use an audio and visual synchronization audio encoder trained on the 2D audio-visual synchronization dataset LRS2~\cite{afouras2018deep}. This encourages the audio features extracted by our method and lip movements to have the same feature distribution. The specific implementation method is as follows: We use a pre-trained lip synchronization discriminator~\cite{chung2017out}. It can give confidence for the lip synchronization effect of the video. The lip synchronization discriminator takes as input a continuous face window $ F $ and the corresponding audio frame $ A $. If they overlap entirely, they are judged as positive samples (with label $ y = 1 $). Otherwise, they are judged as negative samples (with label $ y = 0 $). The discriminator calculates the cosine similarity between these sequences as follows:

\begin{equation}
\text{sim}(F, A) = \frac{F \cdot A}{\|F\|_2 \|A\|_2},
\end{equation}
and then uses binary cross-entropy loss:

\begin{equation}\label{eq:2}
L_{\text{sync}} = -\left( y \log(\text{sim}(F, A)) + (1-y) \log(1-\text{sim}(F, A)) \right),
\end{equation}
to minimize the distance for synchronized samples and maximize the distance for non-synchronized samples.

Under the supervision of the lip synchronization discriminator, we pre-train a highly synchronized audio-visual feature extractor related to lip movements. First, we use convolutional networks to obtain audio features $ \text{Conv}(A) $ and encode facial features $ \text{Conv}(F) $. These features are then concatenated. In the decoding phase, we use stacked convolutional layers to restore facial frames using the operation $\text{Dec}(\text{Conv}(A) \oplus \text{Conv}(F))$. The $L_1$ reconstruction loss during training is given by:   

\begin{equation}
L_{\text{recon}} = \|F - \text{Dec}(\text{Conv}(A) \oplus \text{Conv}(F))\|_1.
\end{equation}

\begin{figure}
\begin{center}
   \includegraphics[width=1.\linewidth]{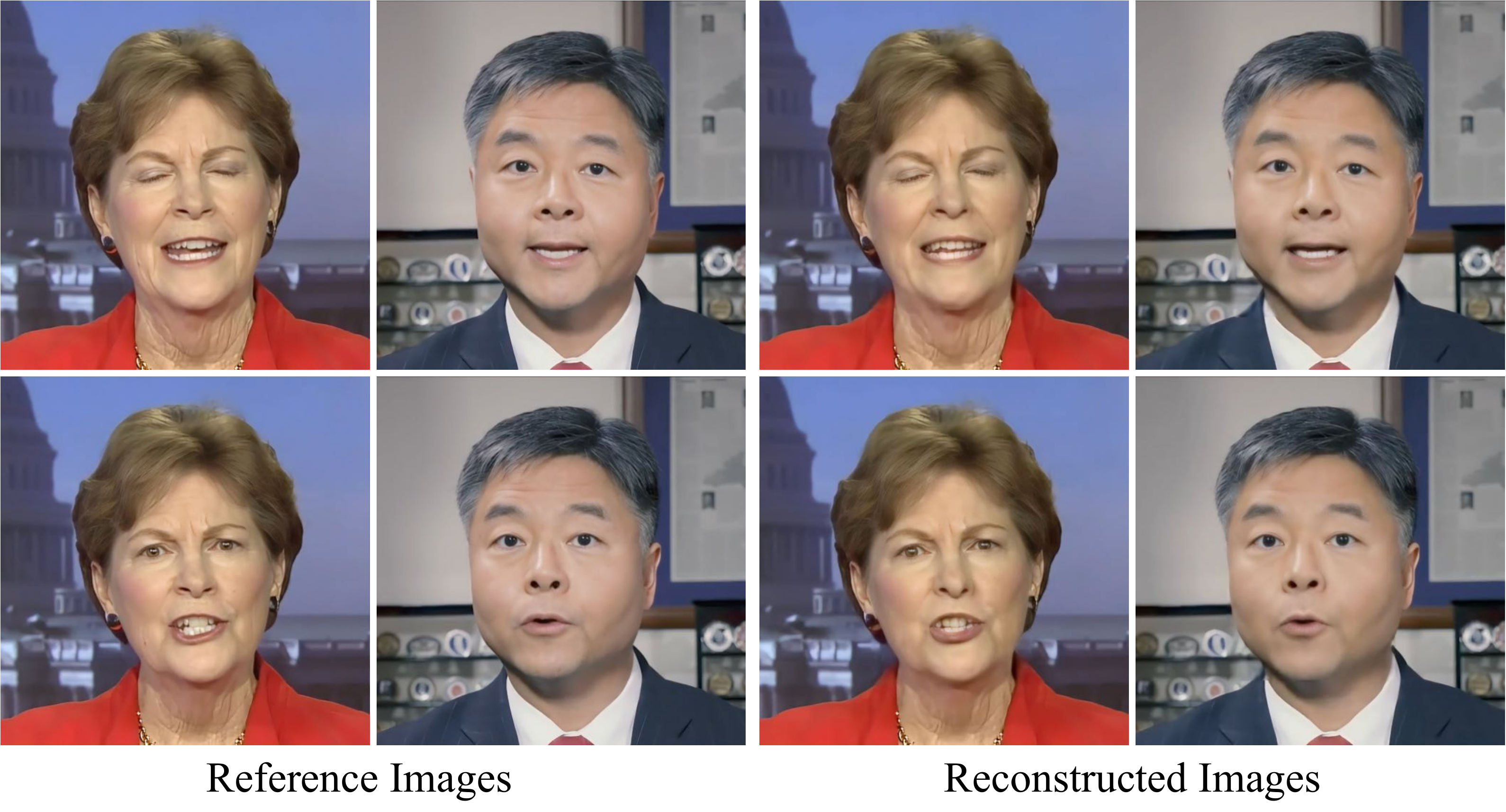}
\end{center}
\vspace{-1em}
   \caption{\textbf{Visualization of reconstruction quality.} The Audio-Visual Encoder effectively captures and reconstructs lip movements.}
\label{fig:recon}
\end{figure}

Simultaneously, we sample synchronized and non-synchronized segments using lip movement discriminators and employ the same sync loss as Eq.~\ref{eq:2}. We train a facial generation network related to audio by minimizing both losses, with the reconstruction results shown in Fig.~\ref{fig:recon}. We discard the facial encoder and decoder parts of the network, retaining only the audio convolution component $ \text{Conv}(A) $, which serves as a highly synchronized audio-visual encoder related to lip movements. Our method effectively restores the lip movements of the input image through audio features, thereby enhancing the lip synchronization capability.

\begin{figure}
\begin{center}
   \includegraphics[width=1.\linewidth]{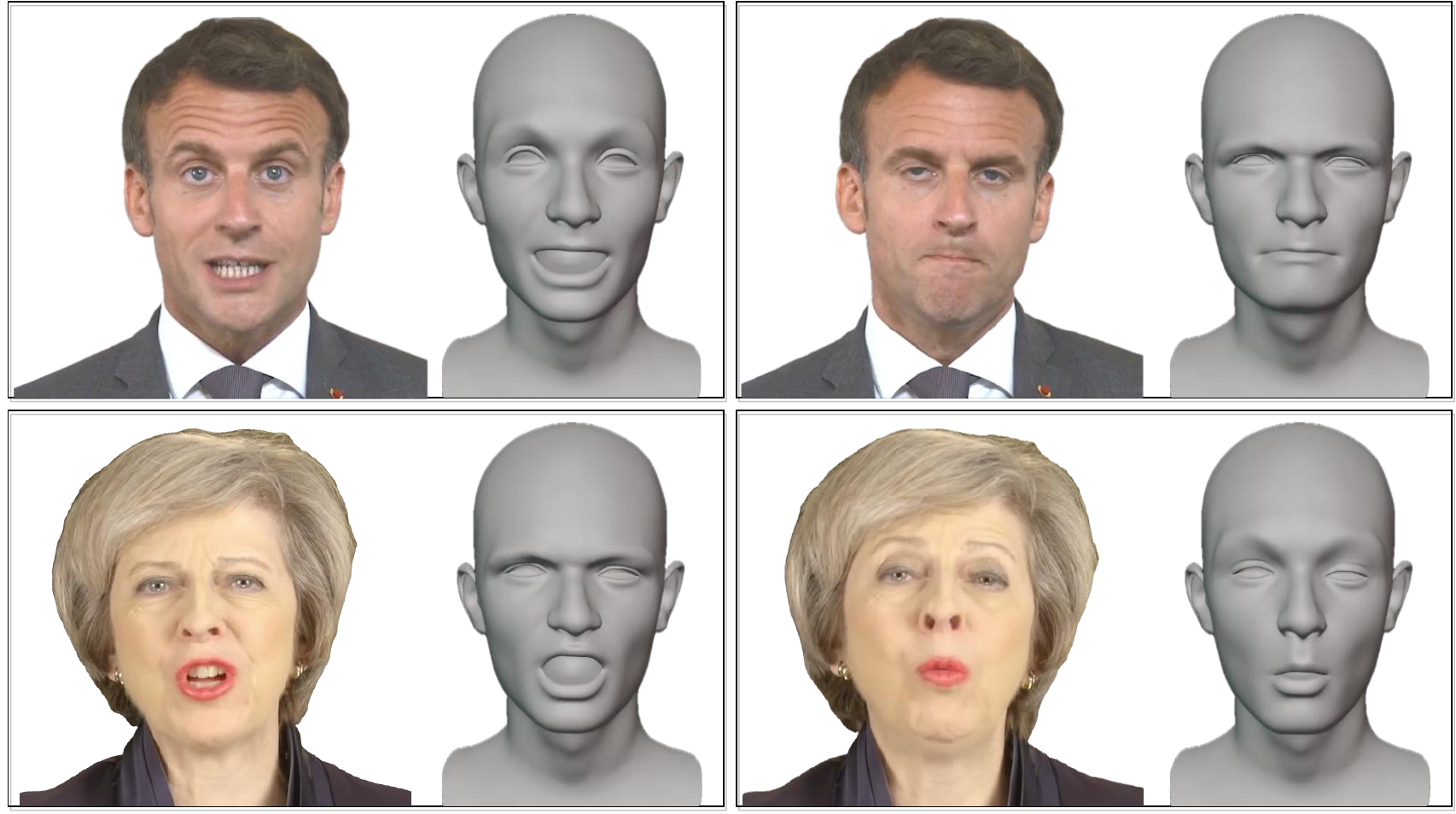}
\end{center}
   \caption{\textbf{Facial Animation Capturer.} We use 3D facial blendshape coefficients to capture the expressions of characters.}
\label{fig:3}
\end{figure}

\noindent\textbf{Facial Animation Capturer.} Considering the need for more synchronized and realistic facial expressions, we add an expression synchronization control module. Specifically, we introduce a 3D facial prior using 52 semantically facial blendshape coefficients~\cite{peng2023emotalk} represented by $ B $ to model the face, as shown in Fig.~\ref{fig:3}. Because the 3D face model can retain the structure information of face motion, it can reflect the content of facial movements well without causing facial structural distortion. During the training process, we first use the facial blendshape capture module, which is composed of ResNet~\cite{he2016deep}, to capture facial expressions as $E(B)$, where $ E $ represents the mapping from the blendshape coefficients to the corresponding facial expression feature. The captured expression can be represented as:

\begin{equation}
E(B) = \sum_{i=1}^{52} w_i \cdot B_i,
\end{equation}
where $ w_i $ are the weights associated with each blendshape coefficient $ B_i $.

We first estimate all 52-dimensional blendshape coefficients and, to facilitate network learning, select seven core facial expression control coefficients—Brow Down Left, Brow Down Right, Brow Inner Up, Brow Outer Up Left, Brow Outer Up Right, Eye Blink Left, and Eye Blink Right—to control the eyebrow, forehead, and eye regions specifically. These coefficients are highly correlated with expressions and are independent of lip movements. The expression of each region can be represented as:

\begin{equation}
E_{\text{core}} = \sum_{j=1}^{7} w_j \cdot B_j,
\end{equation}
where $ E_{\text{core}} $ represents the core expression, $ w_j $ are the corresponding weights for the seven selected blendshape coefficients.

Using these semantically meaningful blendshape coefficients allows the model to capture and accurately represent the nuances of facial movements. During training, this module helps the network learn the complex dynamics of facial expressions more effectively, ensuring that the generated animations maintain structural consistency while being expressive and realistic.

\noindent\textbf{Facial-Aware Masked-Attention.} To reduce the mutual interference between lip features and expression features during training, we use the horizontal coordinate of the nose tip landmark as a boundary to divide the face into two parts: the lower face (lips) and the upper face (expressions). We then apply masks $ M_{\text{lip}} $ and $ M_{\text{exp}} $ to the respective attention areas for lips and expressions. Specifically, the new attention mechanisms are defined as follows:

\begin{equation}
\begin{split}
V_{\text{lip}} &= V \odot M_{\text{lip}}, \\
V_{\text{exp}} &= V \odot M_{\text{exp}}.
\end{split}
\end{equation}

These formulations allow the attention mechanisms to focus solely on their respective parts, thereby reducing entanglement between them. Before the disentanglement, lip movements might induce blinking tendencies and affect hair volume. By introducing the mask module, the attention mechanism can focus on either expressions or lips without affecting other areas, thereby reducing the artifact caused by coupling. Finally, we obtain the disentangled lip feature ${f_l}=f_{\text{lip}} \odot V_{\text{lip}} $ and expression feature ${f_e} = f_{\text{exp}} \odot V_{\text{exp}}$.

\subsection{Head-Sync Stabilizer}
\noindent\textbf{Head Motion Tracker.} The head pose, denoted as $p$, refers to the rotation angle of a person's head in 3D space and is defined by a rotation $R$ and a translation $T$. An unstable head pose can lead to a head jitter. In this section, we use Face Alignment\cite{face-alignment} to extract sparse 2D landmarks and estimate the corresponding 3D keypoints using the BFM (Basel Face Model)\cite{paysan20093d}. The facial shape is modeled with identity ($\alpha_{\text{id}}$) and expression ($\alpha_{\text{exp}}$) parameters, while head motion is captured through rotation ($R$) and translation ($T$). We obtain 3D keypoint projections based on these parameters and compute the projection loss by comparing them with the detected 2D landmarks, allowing for iterative optimization.
For each frame, we refine the expression parameters ($\alpha_{\text{exp}}$), pose parameters ($R, T$), and focal length ($f$), while keeping the identity parameters ($\alpha_{\text{id}}$) fixed to maintain subject consistency. Rather than assuming a rigid 3D facial shape, we explicitly model both static identity features and dynamic expression variations, ensuring robust tracking that captures temporal facial motion changes. Since we do not use expression and identity parameters, we omit them in the following description. The following are the details of the Head Motion Tracker.

Initially, the best focal length is determined through $i$ iterations within a predetermined range. For each focal length candidate, $ f_i $, the system re-initializes the rotation and translation values. The objective is to minimize the error between the projected landmarks from the 3D Morphable Models (3DMM)~\cite{paysan20093d} and the actual landmarks in the video frame.
Formally, the optimal focal length $ f_{\text{opt}} $ is given by:
\begin{equation}
f_{\text{opt}} = \arg\min_{f_i} E_i(L_{2D}, L_{3D}(f_i, R_i, T_i)),
\end{equation}
where $ E_i $ represents the Mean Squared Error (MSE) between these landmarks, $ L_{3D}(f_i, R_i, T_i) $ represents the projected landmarks from the 3DMM for a given focal length $ f_i $, the corresponding rotation and translation parameters $ R_i $ and $ T_i $, $ L_{2D} $ are the actual landmarks from the video frame. 
Subsequently, leveraging the optimal focal length $ f_{\text{opt}} $, the system refines the rotation $ R $ and translation $ T $ parameters for all frames to better align the model's projected landmarks with the actual video landmarks. This refinement process can be mathematically represented as:
\begin{equation}
(R_{\text{opt}}, T_{\text{opt}}) = \arg\min_{R, T} E(L_{2D}, L_{3D}(f_{\text{opt}}, R, T)),
\end{equation}
where $ E $ denotes the MSE metric, between the 3D model's projected landmarks $ L_{3D} $ for the optimal focal length $ f_{\text{opt}} $, and the actual 2D landmarks $ L_{2D} $ in the video frame. The optimized rotation $ R_{\text{opt}} $ and translation $ T_{\text{opt}} $ are obtained by minimizing this error across all frames.

\noindent\textbf{Stable Head Points Tracker.} 
Considering methods based on Gaussian Splatting and their requirements for inputting head rotation $ R $ and translation $ T $, previous methods utilize 3DMM-based techniques to extract head poses and generate an inaccurate result. To improve the precision of $ R $ and $ T $, we use an optical flow estimation model from~\cite{yao2022dfa} to track facial keypoints $ K $. Specifically, we first use a pre-trained optical flow estimation model to obtain optical flow information $ F $ of facial movements. The optical flow information is defined as:
\begin{equation}
F(x_f, y_f, t_f) = (u_f(x_f, y_f, t_f), v_f(x_f, y_f, t_f)),
\end{equation}
where $ u_f $ and $ v_f $ are the horizontal and vertical components of the optical flow at pixel location $(x_f, y_f)$ at time $ t_f $. Then, by applying a Laplacian filter $ L $, we select keypoints with the most significant flow changes:
\begin{equation}
K' = \{ k \in K \mid L(F(k)) > \theta \},
\end{equation}
where $ \theta $ is a threshold defining significant movement. We track these keypoints' movement trajectories $ T_K $ in the optical flow sequence. 

During the optical flow estimation in SyncTalk, we observed noticeable jitter issues when tracking certain subjects, particularly due to the movement of eyebrows and eyes. These regions tend to exhibit more dynamic and unpredictable movements, which can introduce instability in the facial tracking process. To address this, we implemented a Semantic Weighting module that selectively assigns lower weights to key points located in the eyebrow and eye regions, as these are more prone to erratic movements. 

The Semantic Weighting Module first detects sparse landmarks across the face and then applies a semantic weighting mask to the detected keypoints. This step is crucial because the dynamic movements in these regions can otherwise lead to noisy and unstable tracking results. By excluding these high-variance regions, the Semantic Weighting module ensures that only the most stable and reliable keypoints are used in subsequent tracking, significantly enhancing the accuracy of the head pose parameters $ R $ and $ T $.

\noindent\textbf{Bundle Adjustment.} Given the keypoints and the rough head pose, we introduce a two-stage optimization framework from~\cite{guo2021ad} to enhance the accuracy of keypoints and head pose estimations. In the first stage, we randomly initialize the 3D coordinates of $j$ keypoints and optimize their positions to align with the tracked keypoints on the image plane. This process involve minimizing a loss function $ L_{\text{init}} $, which captures the discrepancy between projected keypoints $ P $ and the tracked keypoints $ K'' $, as given by:

\begin{equation} 
L_{\text{init}} = \sum_{j} \lVert P_j - K''_j \rVert_2. 
\end{equation}

Subsequently, in the second stage, we embark on a more comprehensive optimization to refine the 3D keypoints and the associated head jointly pose parameters. Through the Adam Optimization~\cite{kingma2014adam}, the algorithm adjust the spatial coordinates, rotation angles $ R $, and translations $ T $ to minimize the alignment error $ L_{\text{sec}} $, expressed as:

\begin{equation} 
L_{\text{sec}} = \sum_{j} \lVert P_j(R, T) - K''_j \rVert_2. 
\end{equation}

After these optimizations, the resultant head pose and translation parameters are observed to be smooth and stable.

\begin{figure*}
\begin{center}
   \includegraphics[width=1.\linewidth]{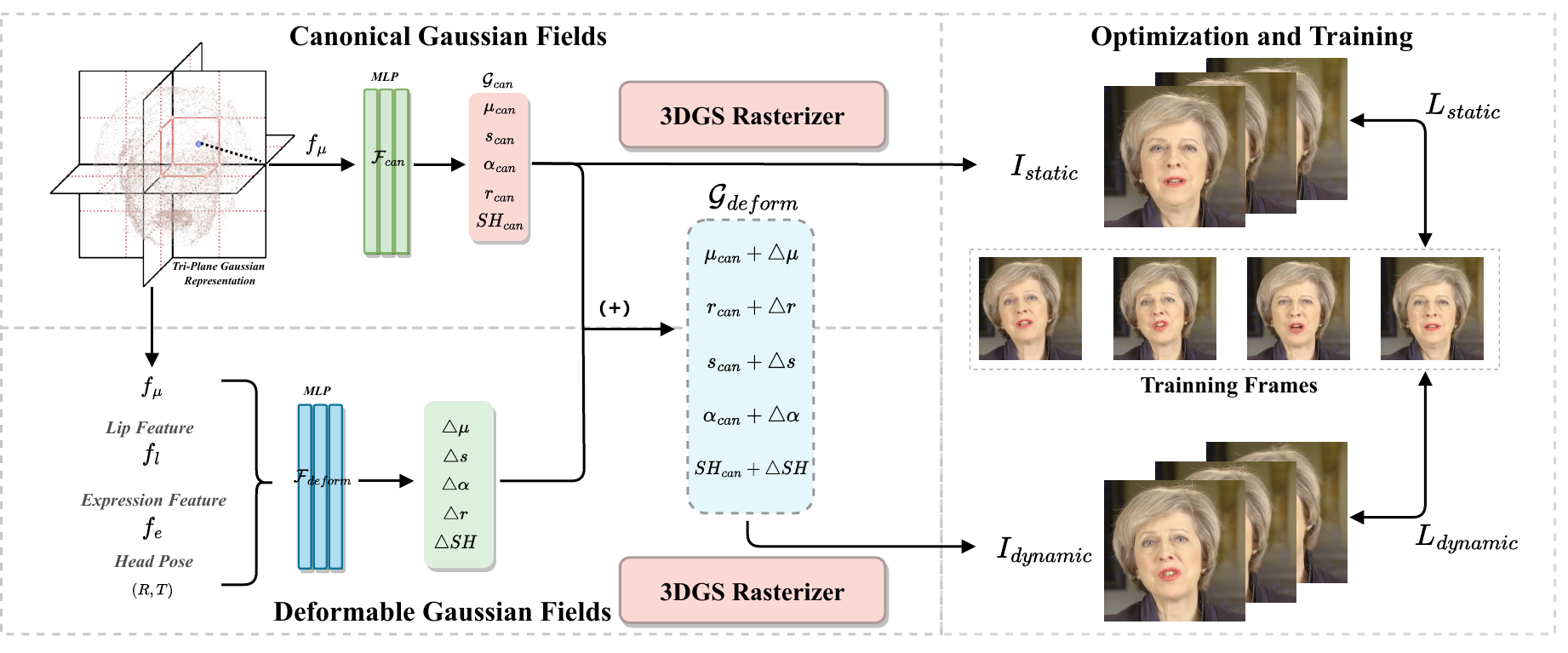}
\end{center}
   \caption{\textbf{Overview of Gaussian Rendering.} Canonical Gaussian fields utilize a triplane representation to encode 3D head features, which are processed by the MLP to yield canonical parameters. These parameters are then integrated with lip feature, expression feature, and head pose parameters in the deformable Gaussian fields. This design facilitates the generation of high-fidelity talking head, achieving realistic and dynamic facial animations.}
\label{fig:gs}
\end{figure*}

\subsection{Dynamic Portrait Renderer}
\noindent\textbf{Preliminaries on 3D Gaussian Splatting.} By leveraging a set of 3D Gaussian primitives and the camera model information from the observational viewpoint, 3D Gaussian Splatting (3DGS)~\cite{kerbl20233d} can be used to calculate the predicted pixel colors. Specifically, each Gaussian primitive can be described by a center mean $\mu \in \mathbb{R}^{3}$ and a covariance matrix $\Sigma \in \mathbb{R}^{3\times 3}$ in the 3D coordinate as follows:
\begin{equation}
\mathit{g}(\mathrm{x})=\exp(-\frac{1}{2}{(\mathrm{x}-\mu)}^{\mathit{T}}{\Sigma }^{-1}(\mathrm{x}-\mu )),
\end{equation}
where the covariance matrix $\Sigma =RS{S}^{T}{R}^{T}$ can be further decomposed into a rotation matrix $R$ and a scaling matrix $S$ for regularizing optimization. These matrices can subsequently be expressed as a learnable quaternion $r\in {\mathbb{R}}^{4}$ and a scaling factor $s\in {\mathbb{R}}^{3}$. For rendering purposes, each Gaussian primitive is characterized by its opacity value $\alpha \in \mathbb{R}$ and spherical harmonics parameters $\mathcal{S}\mathcal{H}\in \mathbb{R}^{k}$, where k is the degrees of freedom. Thus, any Gaussian primitive can be represented as $\mathcal{G}=\{\mu,r,s,\alpha ,\mathcal{S}\mathcal{H}\}$.

During the point-based rendering, the 3D Gaussian is transformed into camera coordinates through the world-to-camera transformation matrix $W$ and projected to image plane via the local affine transformation $J$~\cite{zwicker2001ewa}, such as:
\begin{equation}
{\Sigma }^{\prime}=JW\Sigma {W}^{T}{J}^{T},
\end{equation}

Subsequently, the color of each pixel is computed by blending all the overlapping and depth-sorted Gaussians:

\begin{equation}
\hat{C}(\mathrm{r})=\displaystyle\sum_{i=1}^{N}{c}_{i}\tilde{\alpha _{i}}\displaystyle\prod_{j=1}^{i-1}(1-\tilde{\alpha _{j}}),
\end{equation}
where $i$ is the index of the $N$ Gaussian primitives, ${c}_{i}$ is the view-dependent appearance and $\tilde{\alpha _{i}}$ is calculated from the opacity $\alpha$ of the 3D Gaussian alongside its projected covariance ${\Sigma }^{\prime}$.

\noindent\textbf{Triplane Gaussian Representation.} Utilizing multi-perspective images and corresponding camera poses, we aim to reconstruct canonical 3D Gaussians representing the average shape of a talking head and design a deformation module that modifies these Gaussians based on audio input, as shown in Fig.~\ref{fig:gs}. Ultimately, this deformation module predicts the offset of each Gaussian attribute for the audio input and rasterizes the deformed Gaussians from relevant viewpoints to generate novel images. 

Addressing the challenges of learning canonical 3D Gaussians, such as ensuring consistency across multiple viewpoints, we incorporate three uniquely oriented 2D feature grids~\cite{chan2022efficient,fridovich2023k,hu2023tri}. A coordinate, given by $x=(x,y,z)\in \mathbb{R}^{XYZ}$, undergoes an interpolation process for its projected values via three individual 2D grids:

\begin{equation}
\begin{aligned}
& \text{interp}^{\mathrm{XY}}: (x, y) \rightarrow f^{\mathrm{XY}}(x,y), \\
& \text{interp}^{\mathrm{YZ}}: (y, z) \rightarrow f^{\mathrm{YZ}}(y,z), \\
& \text{interp}^{\mathrm{XZ}}: (x, z) \rightarrow f^{\mathrm{XZ}}(x,z),
\end{aligned}
\end{equation}
where the outputs $f^{\mathrm{XY}}(x,y),f^{\mathrm{YZ}}(y,z),f^{\mathrm{XZ}}(x,z) \in \mathbb{R}^{LD}$, with $L$ representing the number of levels and $D$ representing the feature dimensions per entry, signify the planar geometric features corresponding to the projected coordinates $(x,y),(y,z),(x,z)$. By reducing the dimensionality of the triplane features and projecting them onto the planes based on XYZ coordinates, we can observe that the triplane method effectively models facial depth information while maintaining multi-angle consistency, as shown in Fig.~\ref{fig:xyz}.

By fusing the outcomes, the fused geometric feature $ f_{\mu } \in \mathbb{R}^{3\times LD} $ is derived as:

\begin{figure}
\begin{center}
   \includegraphics[width=1.\linewidth]{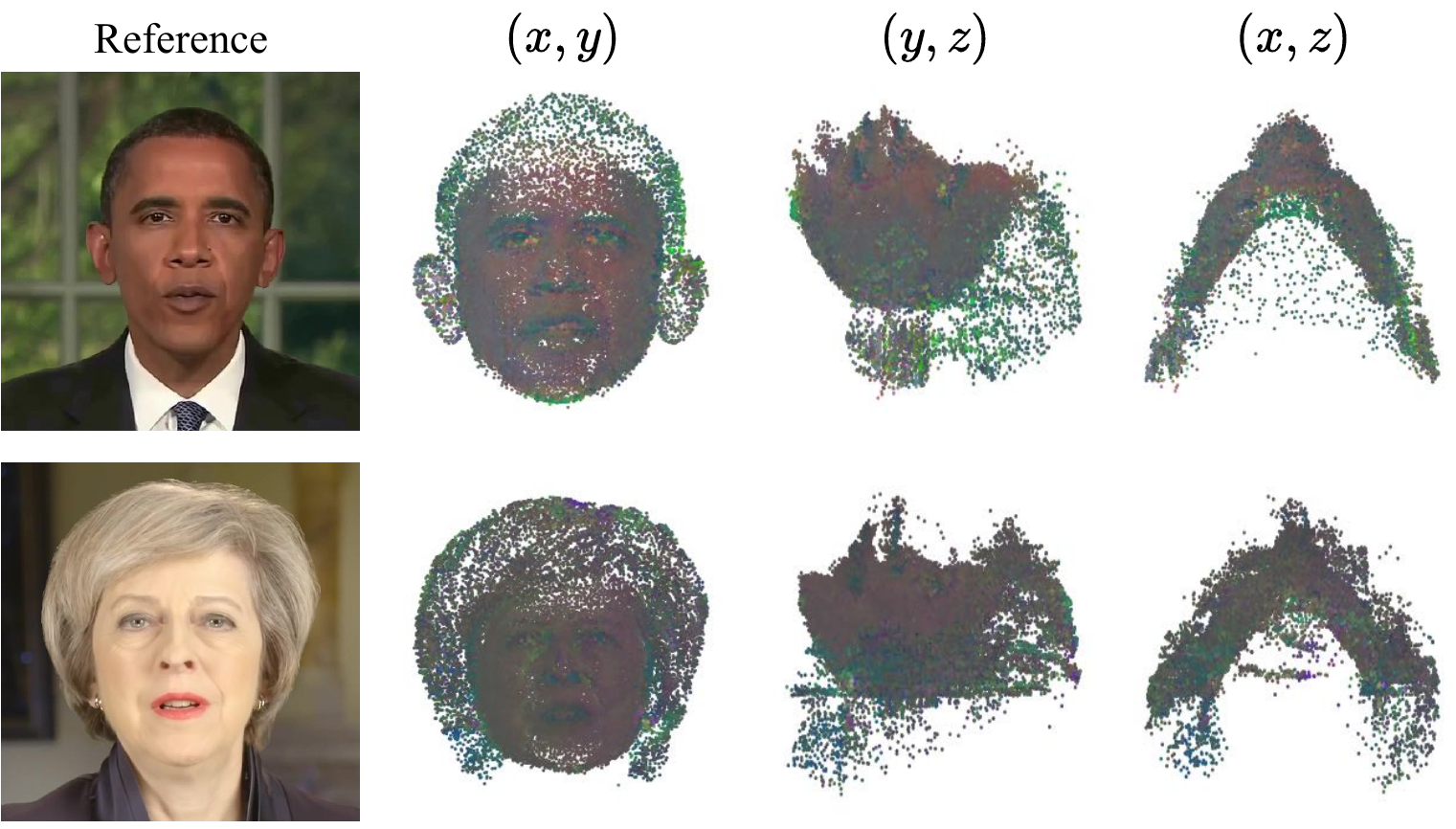}
\end{center}
   \caption{\textbf{Visualization of the triplane feature grids.} The reference images (left) are projected onto three orthogonal planes: $(x, y)$, $(y, z)$, and $(x, z)$.}
\label{fig:xyz}
\end{figure}

\begin{equation}
f_{\mu } = f^{\mathrm{XY}}(x,y) \oplus f^{\mathrm{YZ}}(y,z) \oplus f^{\mathrm{XZ}}(x,z),
\end{equation}
where the concatenation of features is symbolized by $\oplus$, resulting in a $3\times LD$-channel vector. Specifically, we employ a suite of MLP layers, designated as $\mathcal{F}_{\text{can}}$, to project the features $ f_{\mu } $ onto the entire spectrum of attributes of the Gaussian primitives, as illustrated below:

\begin{equation}
{\mathcal{F}}_{\text{can}}\left ( f_{\mu }\right )= {\mathcal{G}}_{\text{can}}=\begin{Bmatrix}
\mu_{c},r_{c},s_{c},\alpha_{c},SH_{c}
\end{Bmatrix}.
\end{equation}

To fully leverage the explicit representation of 3DGS, we opt to deform 3D Gaussians, manipulating not only the appearance information but also the spatial positions and shape of each Gaussian primitive. Consequently, we define a suite of MLP regressors $\mathcal{F}_{\text{deform}}$ to predict the offsets for each Gaussian attribute, utilizing $ f_{\mu } $, the lip feature $ f_l $, and the expression feature $ f_e $, as elucidated below:

\begin{equation}
{\mathcal{F}}_{\text{deform}}\begin{pmatrix}f_{\mu },f_{l},f_{e},R,T
\end{pmatrix}=\begin{Bmatrix}
\triangle \mu,\triangle r,\triangle s,\triangle \alpha,\triangle SH
\end{Bmatrix},
\end{equation}

Thus, by applying the deformation network, we integrate the lip and expression features generated by the Face-Sync Controller module and the head pose features from the Head-Sync Stabilizer module. We then compute the deformations in position, rotation, and scale. These deformations are subsequently integrated with the canonical 3D Gaussians, ultimately defining the deformable 3D Gaussians:
\begin{align}
{\mathcal{G}}_{\text{deform}}=\left\{
\mu_{c}+\triangle \mu,r_{c}+\triangle r,s_{c}+\triangle s,
\notag\right.
\\
\phantom{=\;\;}
\left.\alpha_{c}+\triangle \alpha,SH_{c}+\triangle SH\right\}.
\end{align}

\noindent\textbf{Optimization and Training Details.} We adopt a two-stage training methodology to optimize the model progressively. In the first stage, focused on the canonical Gaussian fields, we begin by optimizing the positions of the 3D Gaussians and the triplanes to establish a preliminary head structure. The static images of the canonical talking head are then rasterized as follows:

\begin{equation}
I_{\text{static}} = \mathcal{R}(\mathcal{G}_{\text{can}}, V),
\end{equation}
where $ V $ defines the camera settings that determine the rendering perspective.

During this stage, we utilize a combination of pixel-level $ \mathcal{L}_{1} $ loss, perceptual loss, and Learned Perceptual Image Patch Similarity (LPIPS) loss to capture fine-grained details and measure the difference between the rendered and real images. The overall loss function is defined as:

\begin{equation}
\begin{split}
\mathcal{L}_{\text{static}} =& \lambda_{\text{L1}}\mathcal{L}_{\text{L1}}+\lambda_{\text{lpips}}\mathcal{L}_{\text{lpips}} +\lambda_{\text{perceptual}}\mathcal{L}_{\text{perceptual}}.
\end{split}
\end{equation}

Once the initial structure is established, we move to the second stage, optimizing the entire network within the deformable Gaussian fields. At this stage, the model predicts deformations as input, and the 3D Gaussian Splatting (3DGS) rasterizer renders the final output images:

\begin{equation}
I_{\text{dynamic}} = \mathcal{R}(\mathcal{G}_{\text{deform}}, V),
\end{equation}
where $ V $ defines the camera settings that determine the rendering perspective.

During the deformation stage, we increase the weight of LPIPS loss, which enhances the model's ability to capture intricate details and textures in the generated images. This focus results in a more realistic and nuanced visual quality compared to the static phase. The loss function used in this stage is:

\begin{equation}
\begin{split}
\mathcal{L}_{\text{dynamic}} =& \lambda_{\text{L1}}\mathcal{L}_{\text{L1}}+\uparrow\lambda_{\text{lpips}}\mathcal{L}_{\text{lpips}}
+\lambda_{\text{perceptual}}\mathcal{L}_{\text{perceptual}}.
\end{split}
\end{equation}

This two-stage approach allows us to refine the model progressively, ensuring structural integrity in the initial phase and high-quality visual output in the final phase. By carefully balancing the various loss terms, we can produce images that are both visually accurate and rich in detail.

\noindent\textbf{Portrait-Sync Generator.}  
To seamlessly blend the 3D Gaussian Splatting (3DGS) rendered facial region with the original high-resolution image while preserving fine details—especially hair strands and subtle textures—we introduce the Portrait-Sync Generator. While 3DGS effectively reconstructs facial structures and expressions, it struggles with high-frequency details such as individual hair strands. This module fuses the 3DGS-rendered facial region $F_r$ with the original high-resolution image $F_o$ (e.g., 1920×1080). Before blending, we apply a Gaussian blur to $F_r$ to generate a smoothed version $G(F_r)$. Then, $G(F_r)$ is placed back onto the original high-resolution image $F_o$ according to the corresponding facial region coordinates. This process enhances the realism of the generated facial region, ensures consistent hair textures across frames, and reduces artifacts, enabling the model to produce high-resolution videos that retain fine details.

\begin{figure}
\begin{center}
   \includegraphics[width=1.\linewidth]{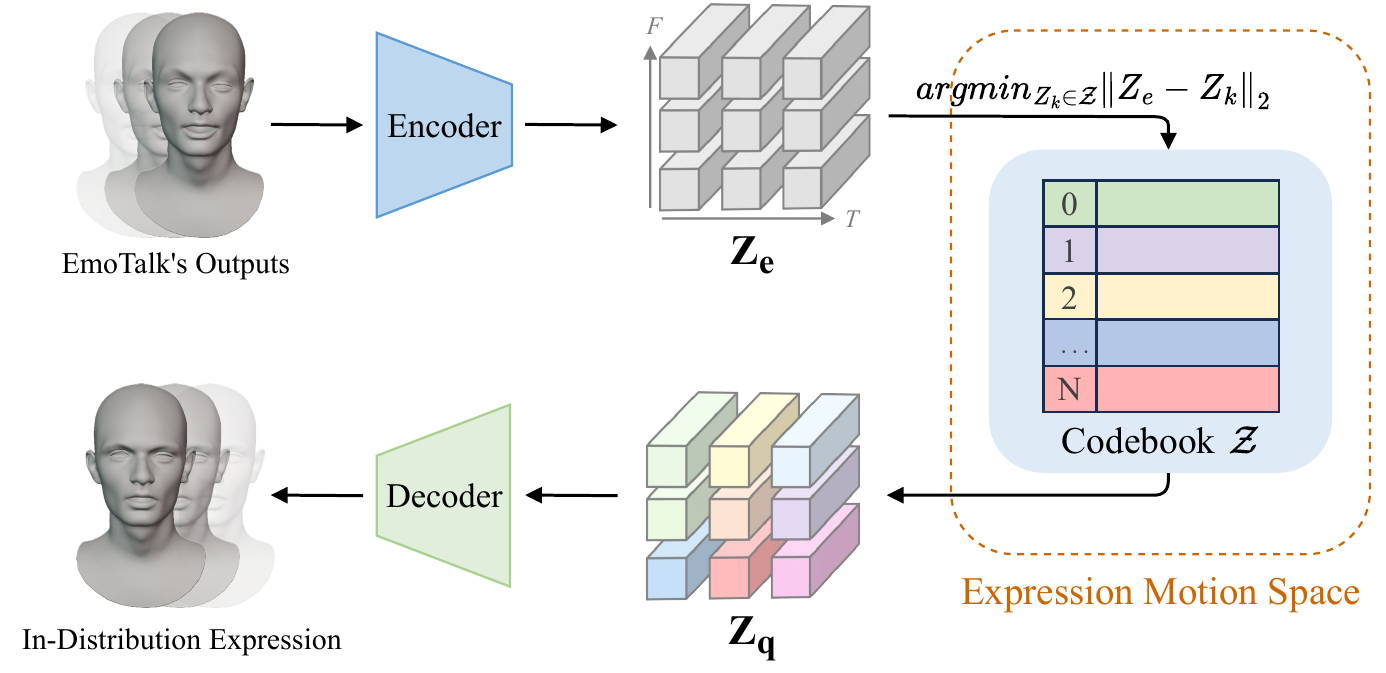}
\end{center}
   \caption{\textbf{Learning framework of blendshape coefficient space.} The VQ-VAE model handles out-of-distribution (OOD) blendshape coefficients by embedding them into a learned codebook, ensuring accurate reconstruction and addressing variations in facial expressions.}
\label{fig:vq}
\end{figure}

\subsection{OOD Audio Expression Generator}

In real-world applications of talking head generation, it is common to encounter scenarios where out-of-distribution (OOD) audio is used. This could include situations where a character’s speech is driven by audio from a different speaker or text-to-speech (TTS) generated audio. However, these situations often lead to a mismatch between the generated facial expressions and the spoken content because previous methods simply repeated facial expressions from the original video. For example, using OOD audio might cause a character to frown while discussing a cheerful topic, thereby undermining the perceived realism and coherence of the generated video.

To overcome these challenges, we introduce an OOD Audio Expression Generator, a module designed to bridge the gap between mismatched audio and facial expression. This generator builds upon our previous work, EmoTalk~\cite{peng2023emotalk}, published at ICCV, which was developed to produce facial expressions that are tightly synchronized with the speech content—what we refer to as speech-matched expressions. EmoTalk provides a more accurate and context-aware method for driving facial expressions based on the audio input, ensuring that the emotional tone and expression match the spoken words.

However, even with EmoTalk, challenges arise when dealing with OOD audio for characters whose facial blendshape coefficients significantly differ from the reference identity used during training. Since renderers based on 3D Gaussian Splatting (3DGS)\cite{kerbl20233d} typically learn facial expression from a limited and specific dataset (e.g., a few-minute-long video), when confronted with OOD blendshape coefficients of cross-identity characters generated by EmoTalk\cite{peng2023emotalk}, the rendering may produce inaccurate facial movements, generate artifacts, or even cause the rendering to crash because 3DGS struggles to extrapolate to unseen expression coefficients from different sources. Therefore, merely improving the quality of blendshape coefficients is insufficient to address the problem of facial expression generation driven by audio of cross-identity characters.

To enhance the generalization ability, we pre-train a Transformer-based VQ-VAE~\cite{van2017neural} model, which includes an encoder $ E $, a decoder $ D $, and a context-rich codebook $ Z $, as shown in Fig.~\ref{fig:vq}. This setup allows the model to effectively capture the characteristic distributions of different identities and generate blendshape coefficients that are tailored to the target character’s facial features during the decoding phase.

Specifically, the encoder $ E $ converts the input blendshape coefficients $ B $ into high-dimensional latent representations $ \mathcal{Z}_e = E(B) $. These representations are then mapped to a discrete embedding vector space using a codebook $ Z = \{z_k \in \mathbb{R}^C\}_{k=1}^N $, where $ C $ represents the dimensionality of each embedding vector, and $ N $ represents the number of codebook entries. The quantization function $ \mathcal{Q} $ maps $ \mathcal{Z}_e $ to its nearest entry in the codebook $ Z $:

\begin{equation}
Z_q = \mathcal{Q}(\mathcal{Z}_e) := \arg\min_{z_k \in Z} \left \| \mathcal{Z}_e - z_k \right \|_2,
\end{equation}
where the quantized embedding vector $ Z_q $ represents the blendshape coefficients adapted for the target character. The reconstructed blendshape coefficients $ \widetilde{B} $ are then generated by the decoder $ D $:

\begin{equation}
\widetilde{B} = D(Z_q) = D(\mathcal{Q}(E(B))).
\end{equation}

This process ensures that the generated blendshape coefficients align with the unique facial features of the target character, even when driven by OOD audio. The discrete codebook helps mitigate mapping ambiguity, allowing the model to retain expressiveness while accurately capturing the discrete features necessary for effective reconstruction.

To supervise the training of the quantized autoencoder, we minimize the reconstruction loss and the quantization loss:

\begin{equation}
\begin{split}
\mathcal{L} = & \ \mathcal{L}_{\text{recon}} + \mathcal{L}_{\text{vq}}  = \left \| B - \widetilde{B} \right \|^2 \\
& + \left \| \mathcal{Z}_e - \text{sg}(Z_q) \right \|^2 + \beta \left \| \text{sg}(\mathcal{Z}_e) - Z_q \right \|^2,
\end{split}
\end{equation}
where $ \text{sg} $ denotes a stop-gradient operation, and $ \beta $ is a weight factor for the commitment loss.

By integrating our method with EmoTalk, we can generate speech-matched facial expressions, even when using OOD audio. The introduction of a discrete codebook enhances the model's ability to generalize across different identities, ensuring that the generated expressions are both consistent and contextually appropriate.

\subsection{OOD Audio Torso Restorer}
Although the Face-Sync Controller, Head-Sync Stabilizer, and Dynamic Portrait Renderer enabled us to achieve high synchronization of facial movements and head poses, challenges remain when it comes to rendering fine textures such as torso, which are distinct from the facial region. Additionally, when generating videos with out-of-distribution (OOD) audio, inconsistencies may arise—such as the character’s mouth is open in the original frame but closed in the generated frame. This discrepancy in jaw position can lead to visible gaps between the generated head and torso, often manifesting as dark areas around the chin.

\begin{figure}
\begin{center}
   \includegraphics[width=1.\linewidth]{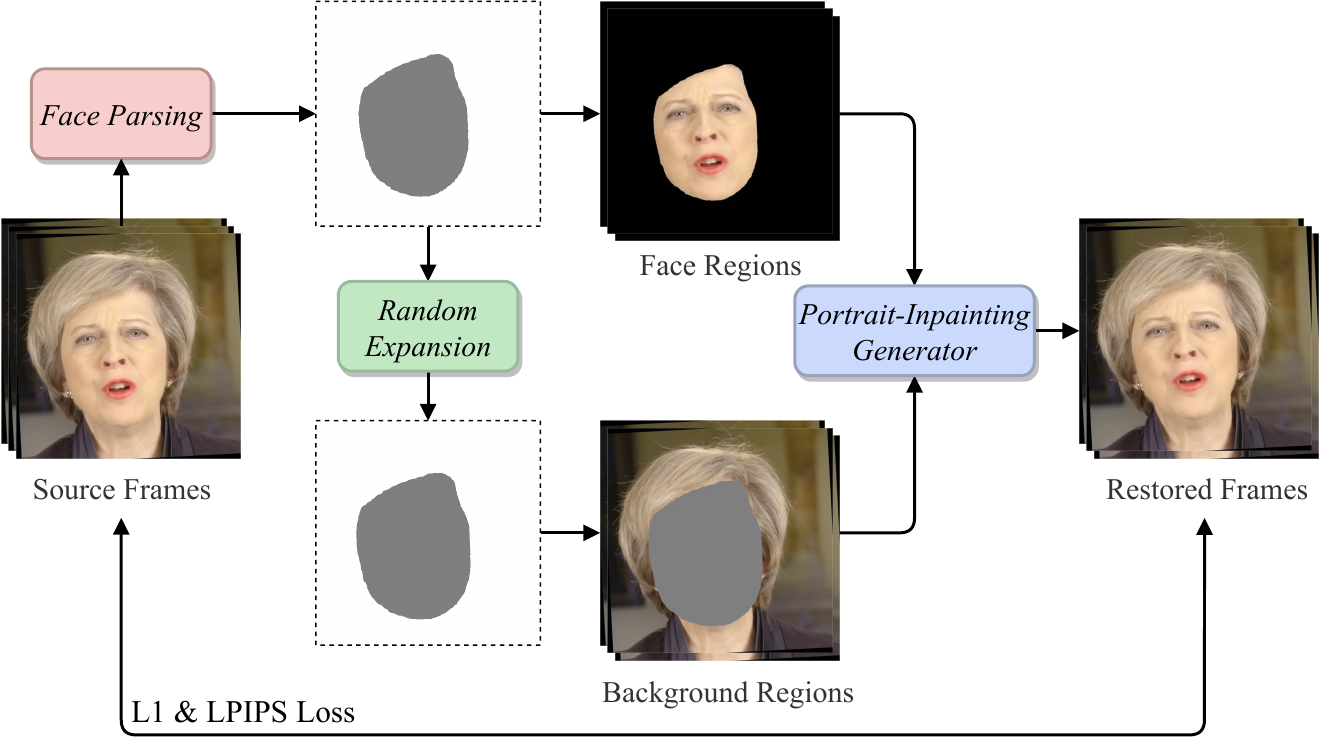}
\end{center}
   \caption{\textbf{Structure of the Torso-Inpainting Restorer.} We manually construct impaired inputs in training to build the network's complementation ability.}
\label{fig:inpaint}
\end{figure}

To address these issues, we develop an OOD Audio Torso Restorer. The main module is the Torso-Inpainting Restorer module, as shown in Fig.~\ref{fig:inpaint}. This module is designed to repair any gaps at the junction between the head and torso due to discrepancies in facial expressions or jaw positions. The Torso-Inpainting utilizes a lightweight U-Net-based inpainting model to seamlessly integrate the rendered facial region with the torso, ensuring the visual coherence and quality of the final output.

The primary cause of these gaps is the mismatch between the facial boundaries rendered by Gaussian Splatting and the torso from the source video. To simulate and address this issue during training, we process the source video frames $ F_{source} $ to obtain the original-sized facial mask $ M $ and the ground truth facial region $ MF_{source} $. To enhance the network's robustness to various poses, we randomly rotate the source frames and expand the cheeks and chin areas of the facial mask. The expanded mask area is then removed from each frame, resulting in a pseudo ground truth for the background region $ (1-M-\delta_{\text{ran}})F_{source} $, where $ \delta_{ran} $ is the random expansion range of the facial mask $ M $.

The inpainting process used by the Torso-Inpainting Restorer is described by the following equation:

\begin{equation}
\mathcal{I}(MF_{source}, (1-M-\delta_{\text{ran}})F_{source}, \theta) = \hat{F}_{source}, 
\end{equation}
where $ \mathcal{I} $ represents the inpainting process, $ \theta $ represents the learnable parameters of $ \mathcal{I} $.

For 512$\times$512-sized images, the random expansion range is set between 10-30 pixels. After concatenating the facial and background regions, they are fed into the inpainting model, which completes and smooths out the areas that were removed due to the random mask expansion in each frame. The reconstruction loss used to optimize $ \mathcal{I} $ is calculated as:

\begin{equation}
\mathcal{L}_{\text{inpaint}} = \mathcal{L}_{\text{L1}}(F_{source}, \hat{F}_{source}) + \mathcal{L}_{\text{LPIPS}}(F_{source}, \hat{F}_{source}),
\end{equation}
where $ \mathcal{L}_{\text{L1}} $ and $ \mathcal{L}_{\text{LPIPS}} $ are the reconstruction and perceptual losses, respectively.

During rendering, a fixed 15-pixel expansion is applied to the facial mask to obtain a robust background region $ (1-M-\delta)F_{source} $. The generated facial region is then smoothly merged with the background region, and the Torso-Inpainting repairs remaining gaps, ensuring the final frames are visually coherent.

\section{Experiments}
\label{sec:experiment}

\subsection{Experimental Settings}
\noindent\textbf{Dataset.} To ensure a fair comparison, we use the same well-edited video sequences from~\cite{guo2021ad,ye2023geneface,li2023efficient},  including English and French. The average length of these videos is approximately 8,843 frames, and each video is recorded at 25 FPS. Except for the video from AD-NeRF~\cite{guo2021ad}, which has a resolution of $450\times450$, all other videos have a resolution of $512\times512$, with the character-centered.

\noindent\textbf{Implementation Process.}
We adopt the same settings as previous work based on NeRF \cite{guo2021ad,ye2023geneface,li2023efficient}. Specifically, we use a few minutes of video of a single subject as training data shot by a static camera. The framework will save $f_l$, $f_e$, and $(R, T)$ as preprocessing steps. During training, the model preloads these data and stores them in memory or on the GPU. In the inference stage, by inputting the audio feature $f_l$, the model can render the character's image and merge the newly generated face with the original image through the pre-saved mask area, ultimately achieving real-time output.

\noindent\textbf{Comparison Baselines.}
For a fair comparison, we re-implement existing methods to conduct reconstruction and synchronization experiments, including 2D generation-based methods: Wav2Lip \cite{prajwal2020lip}, VideoReTalking \cite{cheng2022videoretalking}, DINet \cite{zhang2023dinet}, TalkLip \cite{wang2023seeing}, IP-LAP \cite{zhong2023identity}, and 3D reconstruction-based methods: AD-NeRF \cite{guo2021ad}, RAD-NeRF \cite{tang2022real}, GeneFace \cite{ye2023geneface}, ER-NeRF \cite{li2023efficient}, TalkingGaussian \cite{li2024talkinggaussian}, and GaussianTalker~\cite{cho2024gaussiantalker}. 

In the head reconstruction experiment, we input the original audio to reconstruct speaking head videos. Taking a subject named ``May'' as an example, we crop the last 553 frames as the test set for the reconstruction experiment and the corresponding audio as the input for inference. In the 2D generation-based methods, we use the officially provided pre-trained models for inference, with video streams input at 25 FPS and the corresponding audio, resulting in the respective outcomes after processing by these five methods. In the 3D reconstruction-based methods, since AD-NeRF \cite{guo2021ad}, RAD-NeRF \cite{tang2022real}, ER-NeRF \cite{li2023efficient}, TalkingGaussian~\cite{li2024talkinggaussian}, and GaussianTalker~\cite{cho2024gaussiantalker} do not provide pre-trained models for the corresponding subjects, we re-train the models for these subjects following the publicly available code. The dataset is divided in the same manner as before methods, and the test results are obtained.

In the synchronization experiment, we choose speeches from other people as the input audio. We test recent methods using the same test sequence as in the head reconstruction experiment, with audio inputs for ER-NeRF \cite{li2023efficient} using the same OOD audio. After obtaining the synthesized video sequences, we use the same evaluation code as Wav2Lip \cite{prajwal2020lip} for assessment, finally obtaining metrics on lip synchronization performance for different methods.

\begin{table*}[]
\setlength\tabcolsep{5pt}
\begin{center}
\resizebox{\linewidth}{!}{
\begin{tabular}{@{}c@{\hspace{6pt}}lcccc|cc|ccc|cc@{}} 
\toprule
\multicolumn{2}{l}{Methods}                     & PSNR ↑           & LPIPS ↓         & MS-SSIM ↑       & FID ↓           & NIQE ↓           & BRISQUE ↓        & LMD ↓           & AUE ↓           & LSE-C ↑    & Time ↓ & FPS ↑     \\ \midrule
\multirow{6}{*}{\rotatebox[origin=c]{90}{2D Generation}} & Wav2Lip \mysmall{(ACM MM 20 \cite{prajwal2020lip})} & 33.4385          & 0.0697          & 0.9781          & 16.0228         & 14.5367          & 44.2659          & 4.9630          & 2.9029          & \textbf{9.2387}  & - & 21.26 \\
& \begin{tabular}[c]{@{}l@{}}VideoReTalking\\\mysmall{(SIGGRAPH Asia 22 \cite{cheng2022videoretalking})}\end{tabular}   & 31.7923          & 0.0488          & 0.9680          & 9.2063         & 14.2410          & 43.0465          & 5.8575          & 3.3308          & 7.9683    & -  & 0.76     \\
& DINet \mysmall{(AAAI 23 \cite{zhang2023dinet})}     & 31.6475          & 0.0443          & 0.9640          & 9.4300          & 14.6850          & 40.3650          & 4.3725          & 3.6875          & 6.5653  & -   & 23.74      \\
& TalkLip \mysmall{(CVPR 23 \cite{wang2023seeing})}   & 32.5154          & 0.0782          & 0.9697          & 18.4997         & 14.6385          & 46.6717          & 5.8605          & 2.9579          & 5.9472   & -  & 3.41      \\
& IP-LAP \mysmall{(CVPR 23 \cite{zhong2023identity})}    & 35.1525          & 0.0443          & {\ul 0.9803}    & 8.2125          & 14.6400          & 42.0750      & 3.3350    & 2.8400   & 4.9541   & -   &3.27               \\ \midrule
\multirow{10}{*}{\rotatebox[origin=c]{90}{3D Reconstruction}} & AD-NeRF \mysmall{(ICCV 21 \cite{guo2021ad})}   & 26.7291          & 0.1536          & 0.9111          & 28.9862        & 14.9091          & 55.4667          & 2.9995          & 5.5481          & 4.4996  & 16.4h   & 0.14      \\
& RAD-NeRF \mysmall{(arXiv 22 \cite{tang2022real})}    & 31.7754          & 0.0778          & 0.9452          & 8.6570          & {\ul 13.4433}    & 44.6892          & 2.9115          & 5.0958          & 5.5219  & 5.2h   & 53.87      \\
& GeneFace \mysmall{(ICLR 23 \cite{ye2023geneface})}  & 24.8165          & 0.1178          & 0.8753          & 21.7084         & 13.3353          & 46.5061          & 4.2859          & 5.4527          & 5.1950  & 12.3h   & 7.79      \\
& ER-NeRF \mysmall{(ICCV 23 \cite{li2023efficient})}   & 32.5216          & 0.0334          & 0.9501          & 5.2936          & 13.7048          & {\ul 34.7361}    & 2.8137          & 4.1873          & 5.7749 & 3.1h  & 55.41        \\ 
& \begin{tabular}[c]{@{}l@{}}TalkingGaussian\\\mysmall{(ECCV 24 \cite{li2024talkinggaussian})}\end{tabular}	&33.1178	&0.0333	&0.9610	&7.5019	&13.9055	&41.6166		&2.6974	&2.6686 &6.1842 & \textbf{1.2h}  & \textbf{105.36} \\
& \begin{tabular}[c]{@{}l@{}}GaussianTalker\\\mysmall{(ACM MM 24 \cite{cho2024gaussiantalker})}\end{tabular}	&32.7476	&0.0523		&0.9541	&6.3916	&14.0236	&43.2154		&2.8308	&4.0350	&6.2074 & 2.8h & 85.03 \\

\cmidrule(l){2-13} 
& SyncTalk \mysmall{(CVPR 24 \cite{peng2024synctalk})}     & {\ul 35.3542}    & {\ul 0.0235}    & 0.9769          & {\ul 3.9247}    & \textbf{13.1333} & \textbf{33.2954} & {\ul 2.5714}    & {\ul2.5796} & {\ul 8.1331} & 3.2h   & 52.18  \\
& SyncTalk++ (Ours)    & \textbf{36.3779}	&\textbf{0.0201} &\textbf{0.9826}	&\textbf{3.8771}	&13.7215	&39.3769		&\textbf{2.5331}	&\textbf{2.5211} &7.8298 & {\ul 1.5h}  & {\ul 101.20}  \\ \bottomrule
\end{tabular}}
\end{center}
\caption{\textbf{Quantitative results of head reconstruction.} We achieve state-of-the-art performance on most metrics. We highlight \textbf{best} and \underline{second-best} results.}
\label{tab:1}
\end{table*}

\begin{table}[]
\begin{center}

\begin{tabular}{@{}lcccc@{}}
\toprule
                           & PSNR ↑           & LPIPS ↓         & MS-SSIM ↑       & FID ↓           \\ \midrule
SyncTalk (w/o Portrait)    & 35.3542          & 0.0235          & 0.9768          & 3.9247          \\
SyncTalk (Portrait)      & {\ul 37.4016}    & {\ul 0.0113}    & {\ul 0.9841}    & {\ul 2.7070}    \\
SyncTalk++  (w/o Portrait) & 36.3779          & 0.0201          & 0.9826          & 3.8771          \\
SyncTalk++ (Portrait)    & \textbf{39.5748} & \textbf{0.0097} & \textbf{0.9905} & \textbf{2.1958} \\ \bottomrule
\end{tabular}
\end{center}
\caption{\textbf{Quantified Results of Portrait Mode.} ``Portrait'' refers to the use of the Portrait-Sync Generator. SyncTalk++ outperforms SyncTalk on all metrics.}
\label{tab:protrait}
\end{table}

\begin{table}[]
\begin{center}
\resizebox{\linewidth}{!}{
\begin{tabular}{@{}lcccc@{}}
\toprule
\multicolumn{1}{c}{\multirow{2.5}{*}{Methods}} & \multicolumn{2}{c}{Audio A}       & \multicolumn{2}{c}{Audio B}       \\ \cmidrule(l){2-5} 
\multicolumn{1}{c}{}                         & LSE-D ↓         & LSE-C ↑         & LSE-D ↓         & LSE-C ↑         \\ \midrule

DINet \mysmall{(AAAI 23 \cite{zhang2023dinet})}                                        & {\ul 8.5031}          & 5.6956          & {\ul 8.2038}          & 5.1134          \\
TalkLip \mysmall{(CVPR 23 \cite{wang2023seeing})}                                      & 8.7615          & {\ul 5.7449}          & 8.7019          & {\ul 5.5359}          \\
IP-LAP \mysmall{(CVPR 23 \cite{zhong2023identity})}                                       & 9.8037          & 3.8578          & 9.1102          & 4.389           \\
GeneFace \mysmall{(ICLR 23 \cite{ye2023geneface})}                                     & 9.5451          & 4.2933          & 9.6675          & 3.7342          \\
ER-NeRF \mysmall{(ICCV 23 \cite{li2023efficient})}                                      & 11.813          & 2.4076          & 10.7338         & 3.0242          \\
\multirow{2}{*}{\shortstack[l]{TalkingGaussian \\ \mysmall{(ECCV 24 \cite{li2024talkinggaussian})}}} & \multirow{2}{*}{9.3027} & \multirow{2}{*}{4.8452} & \multirow{2}{*}{9.699} & \multirow{2}{*}{4.2032} \\
& & & & \\
\multirow{2}{*}{\shortstack[l]{GaussianTalker \\ \mysmall{(ACM MM 24 \cite{cho2024gaussiantalker})}}} & \multirow{2}{*}{10.1228} & \multirow{2}{*}{4.2625} & \multirow{2}{*}{10.0872} & \multirow{2}{*}{3.8152} \\
& & & & \\
SyncTalk++ (Ours)                                     & \textbf{8.0808}    & \textbf{6.4633}    & \textbf{8.0217} & \textbf{6.0733} \\ \bottomrule
\end{tabular}}
\end{center}

\caption{\textbf{Quantitative results of the lip synchronization.} We use two different audio samples to drive the same subject, then highlight \textbf{best} and \underline{second-best} results.}
\label{tab:2}
\end{table}

\begin{table}[]
\begin{center}
\resizebox{\linewidth}{!}{
\begin{tabular}{@{}lccccc@{}}
\toprule
Method      & PSNR ↑  & LPIPS ↓   & LMD ↓ & LSE-C ↑ \\ \midrule
$s,r,\mathcal{S}\mathcal{H},\alpha$    & 31.468 & 0.053  & 2.799 & 8.904 \\
$\mathcal{S}\mathcal{H},\alpha$        & \textbf{33.759} & \textbf{0.032}   & \textbf{2.772} & \textbf{9.373} \\
$r,s$         & 30.905 & 0.071  & 2.849 & 8.307 \\
random init & 29.902 & 0.089  & 2.905 & 8.105 \\ \bottomrule
\end{tabular}}
\end{center}
\caption{\textbf{Result of Different Initialization Strategies on 3D Head Representation.} We evaluate the impact of various initialization strategies on facial reconstruction quality, demonstrating their effects on synchronization and visual fidelity.}
\label{tab:canonical}
\end{table}

\subsection{Quantitative Evaluation}
\noindent\textbf{Full Reference Quality Assessment.} In terms of image quality, we use full reference metrics such as Peak Signal-to-Noise Ratio (PSNR), Learned Perceptual Image Patch Similarity (LPIPS)~\cite{zhang2018unreasonable}, Multi-Scale Structure Similarity (MS-SSIM), and Frechet Inception Distance (FID)~\cite{heusel2017gans} as evaluation metrics. 

\noindent\textbf{No Reference Quality Assessment.} In high PSNR images, texture details may not align with human visual perception~\cite{zhang2019ranksrgan}. For more precise output definition and comparison, we use three No Reference methods: the Natural Image Quality Evaluator (NIQE)\cite{mittal2012making}, the Blind/Referenceless Image Spatial Quality Evaluator (BRISQUE)~\cite{mittal2012no} and the Blindly Assess Image Quality By Hyper Network(HyperIQA)~\cite{Su_2020_CVPR}.

\noindent\textbf{Synchronization Assessment.} For synchronization, we use landmark distance (LMD) to measure the synchronicity of facial movements, action units error (AUE)~\cite{baltruvsaitis2015cross} to assess the accuracy of facial movements, and introduce Lip Sync Error Confidence (LSE-C), consistent with Wav2Lip~\cite{prajwal2020lip}, to evaluate the synchronization between lip movements and audio.

\noindent\textbf{Efficiency Assessment.} To evaluate the computational efficiency of our model, we measure both training time and inference speed. Training time reflects the total duration required for the model to converge on a given dataset. For real-time applicability, we assess inference speed in terms of frames per second (FPS) during video generation, where a higher FPS indicates better real-time performance, making the model more suitable for applications such as live streaming and video conferencing.

\noindent\textbf{Evaluation Results.} The evaluation results of the head reconstruction are shown in Tab.\ref{tab:1}. We compare the latest methods based on 2D generation and 3D reconstruction. It can be observed that our image quality is superior to other methods in all aspects. Because we can maintain the subject’s identity well, we surpass 2D generation-based methods in image quality. Due to the synchronization of lips, expressions, and poses, we also outperform 3D reconstruction-based methods in image quality. Particularly in terms of the LPIPS metric, our method has a 65.67$\%$ lower error compared to the previous state-of-the-art method, TalkingGaussian~\cite{li2024talkinggaussian}. In terms of lip synchronization, our results surpass most methods, proving the effectiveness of our Audio-Visual Encoder. We also compare the two output modes of SyncTalk++, one processed through the Portrait-Sync Generator and one without, as shown in Tab.\ref{tab:protrait}. After processing through the Portrait-Sync Generator, hair details are restored, and image quality is improved. Compared with SyncTalk, SyncTalk++ shows significantly better image quality, demonstrating the robustness of our introduction of Gaussian splatting for rendering. We compare the latest SOTA method drivers using out-of-distribution (OOD) audio, and the results are shown in Tab.\ref{tab:2}. We introduce Lip Synchronization Error Distance (LSE-D) and Confidence (LSE-C) for lip-speech sync evaluation, aligning with\cite{prajwal2020lip}. Our method shows state-of-the-art lip synchronization, overcoming small-sample 3D reconstruction limitations by incorporating a pre-trained audio-visual encoder for lip modeling.

We also evaluate the training time and rendering speed. On an NVIDIA RTX 4090 GPU, our method requires only 1.5 hours to train for a new character and achieves 101 FPS at a resolution of $512 \times 512$, far exceeding the 25 FPS video input speed, enabling real-time video stream generation. Compared to SyncTalk~\cite{peng2024synctalk}, SyncTalk++ achieves a shorter training time and higher rendering speed.

\noindent\textbf{Impact of Initialization Strategies on Canonical 3D Head Representation.} To assess the effectiveness of our approach, we compare different initialization strategies for the canonical 3D head representation. As shown in Tab.~\ref{tab:canonical}, the $\mathcal{SH}, \alpha$ initialization achieves the best overall performance, leading to higher image quality and synchronization accuracy.

Compared to other methods, $\mathcal{SH}, \alpha$ results in lower LPIPS and LMD scores, indicating improved perceptual quality and facial alignment. This suggests that leveraging spherical harmonics ($\mathcal{SH}$) and opacity ($\alpha$) attributes effectively enhances spatial consistency and feature learning. In contrast, random initialization leads to degraded performance, highlighting the importance of structured attribute conditioning.

Interestingly, using all attributes ($s, r, \mathcal{SH}, \alpha$) does not yield the best results. This is likely because introducing too many attributes increases optimization complexity and potential redundancy, which can make it harder for the model to focus on the most critical features for synchronization and reconstruction. Given these results, we adopt $\mathcal{SH}, \alpha$ as our default initialization strategy, as it offers the best trade-off between visual fidelity and synchronization accuracy.
\begin{figure*}
\begin{center}
   \includegraphics[width=1.\linewidth]{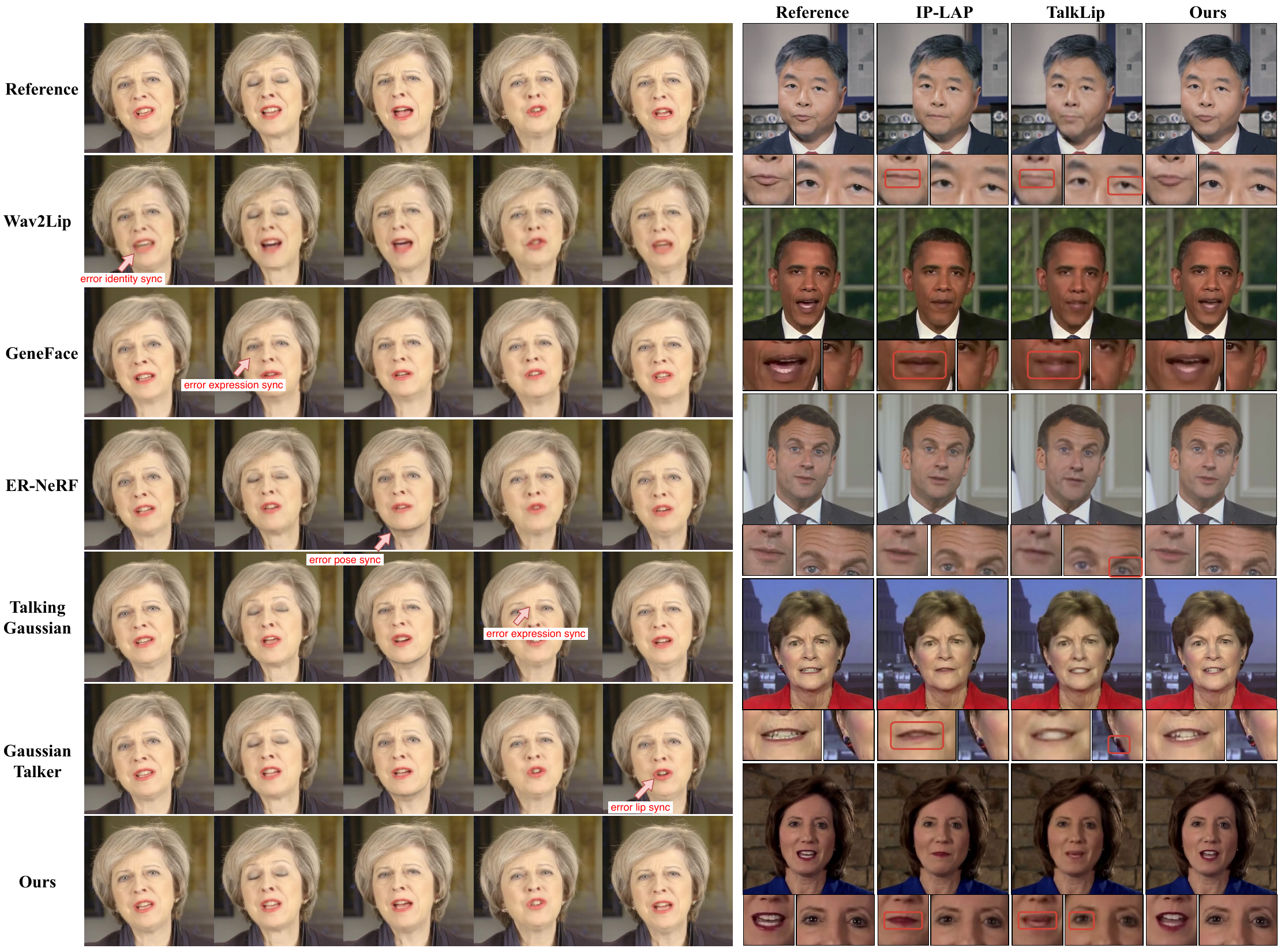}
\end{center}
   \caption{\textbf{Qualitative comparison of facial synthesis by different methods.} Our method has the best visual effect on lip movements and facial expressions without the problem of separation of head and torso. Please zoom in for better visualization.}
\label{fig:compare}
\end{figure*}

\begin{figure}
\begin{center}
\includegraphics[width=1.\linewidth]{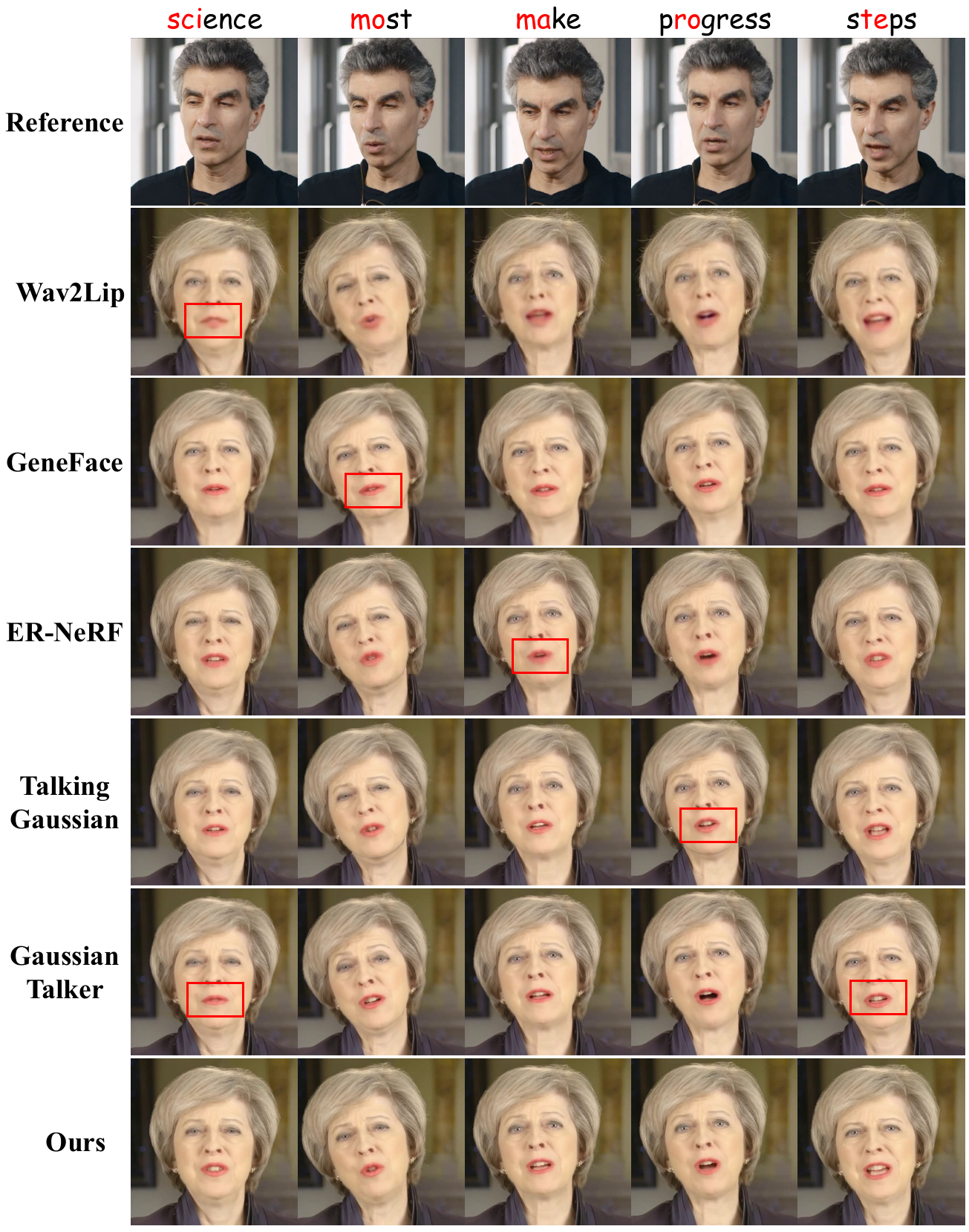}
\end{center}
   \caption{\textbf{Qualitative comparison of facial synthesis driven by in-the-wild audios.}} Our method demonstrates the most accurate lip movement while maintaining the subject's identity well.
\label{fig:ood}
\end{figure}

\subsection{Qualitative Evaluation}
\noindent\textbf{Evaluation Results.} 
To more intuitively evaluate image quality, we display a comparison between our method and other methods in Fig.~\ref{fig:compare}. In this figure, it can be observed that SyncTalk++ demonstrates more precise and more accurate facial details. Compared to Wav2Lip~\cite{prajwal2020lip}, our method better preserves the subject's identity while offering higher fidelity and resolution. Against IP-LAP~\cite{zhong2023identity}, our method excels in lip shape synchronization, primarily due to the audio-visual consistency brought by the audio-visual encoder. Compared to GeneFace~\cite{ye2023geneface}, our method can accurately reproduce actions such as blinking and eyebrow-raising through expression sync. In contrast to ER-NeRF~\cite{li2023efficient}, our method avoids the separation between the head and body through the Pose-Sync Stabilizer and generates more accurate lip shapes. Our method achieves the best overall visual effect; we recommend watching the supplementary video for comparison.

\begin{table}[]
\setlength\tabcolsep{2pt}
\begin{center}
\resizebox{\linewidth}{!}{
\begin{tabular}{@{}lccccc@{}}
\toprule
                Methods& \begin{tabular}[c]{@{}c@{}}Lip-sync\\ Accuracy\end{tabular} & \begin{tabular}[c]{@{}c@{}}Exp-sync\\ Accuracy\end{tabular} & \begin{tabular}[c]{@{}c@{}}Pose-sync\\ Accuracy\end{tabular} & \begin{tabular}[c]{@{}c@{}}Image\\ Quality\end{tabular} & \begin{tabular}[c]{@{}c@{}}Video\\ Realness\end{tabular} \\ \midrule
Wav2Lip\mysmall{~\cite{prajwal2020lip}}         & 4.029             & 3.531             & 3.571              & 3.183          & 3.371          \\
VideoReTalking\mysmall{~\cite{cheng2022videoretalking}}  & 3.594             & 3.606             & 3.771              & 3.446          & 3.537          \\
DINet\mysmall{~\cite{zhang2023dinet}}           & 3.349             & 3.474             & 3.520               & 3.234          & 3.337          \\
TalkLip\mysmall{~\cite{wang2023seeing}}         & 3.657             & 3.554             & 3.874              & 3.451          & 3.491          \\
IP-LAP\mysmall{~\cite{zhong2023identity}}          & 3.943             & {\ul 4.063}       & 4.103              & 4.051          & 4.046          \\ \midrule
AD-NeRF\mysmall{~\cite{guo2021ad}}         & 3.360              & 3.497             & 3.286              & 3.343          & 3.303          \\
RAD-NeRF\mysmall{~\cite{tang2022real}}        & 3.406             & 3.583             & 3.320               & 3.440           & 3.354          \\
GeneFace\mysmall{~\cite{ye2023geneface}}        & 3.086             & 3.389             & 3.166              & 3.257          & 3.189          \\
ER-NeRF\mysmall{~\cite{li2023efficient}}         & 3.640              & 3.634             & 3.526              & 3.617          & 3.674          \\
GaussianTalker\mysmall{~\cite{cho2024gaussiantalker}}  & 3.857             & 3.926             & 3.909              & 3.994          & 4.011          \\
TalkingGaussian\mysmall{~\cite{li2024talkinggaussian}} & 3.703             & 3.731             & 3.691              & 3.800            & 3.903          \\ \midrule
SyncTalk\mysmall{~\cite{peng2024synctalk}}        & {\ul 4.131}       & 4.034             & {\ul 4.149}        & {\ul 4.069}    & {\ul 4.097}    \\
SyncTalk++ (Ours)      & \textbf{4.309}    & \textbf{4.154}    & \textbf{4.371}     & \textbf{4.297} & \textbf{4.229} \\ \bottomrule
\end{tabular}
}
\end{center}
\caption{\textbf{User Study.} Rating is on a scale of 1-5; the higher, the better. The term ``Exp-sync Accuracy'' is an abbreviation for ``Expression-sync Accuracy''. We highlight \textbf{best} and \underline{second-best} results.}
\label{tab:User}
\end{table}

To comprehensively evaluate the method's performance in real-world scenarios, as shown in Fig.~\ref{fig:ood}, we present a qualitative comparison of lip-sync effects driven by in-the-wild audios. The Wav2Lip~\cite{prajwal2020lip}, while producing relatively realistic facial animations, exhibits significant discrepancies in lip-audio synchronization, such as misalignment during the pronunciation of ``science.'' GeneFace~\cite{ye2023geneface} shows some improvement, but synchronization remains unnatural on key syllables. ER-NeRF~\cite{li2023efficient} enhances lip-sync performance; however, during the pronunciation of ``make,'' the lip movements do not fully match the audio. 
Talking Gaussian~\cite{li2024talkinggaussian} produces realistic results with detailed facial handling, but lip movements still show discrepancies. During ``progress,'' lip-audio synchronization is poor, with noticeable lag. Gaussian Talker~\cite{cho2024gaussiantalker} offers more consistent lip-sync but shows rigidity during fast syllable transitions and struggles with complex syllables, resulting in less natural lip movements. In contrast, our method generates superior lip-sync effects driven by in-the-wild audios, demonstrating higher reliability and naturalness in both coherence and detail accuracy. This indicates our method excels at capturing and reproducing complex lip movements in the wild audios, enhancing lip-sync quality and achieving optimal visual effects.

\begin{figure}
\begin{center}
   \includegraphics[width=1\linewidth]{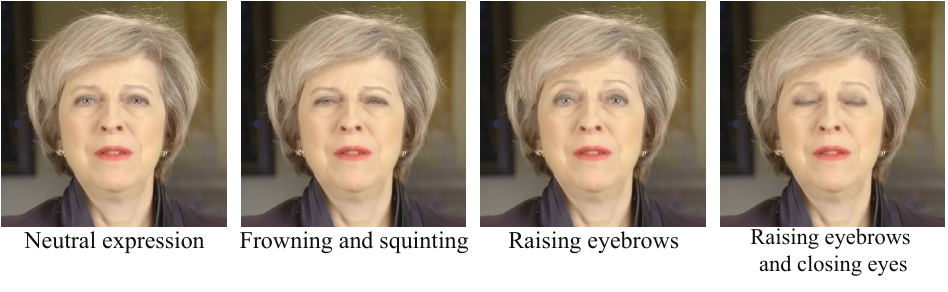}
\end{center}
   \caption{\textbf{Expression generation using OOD Audio Expression Generator.} For different expression coefficients, our method can achieve highly accurate eyebrow and eye generation.}
\label{fig:bs}
\end{figure}

Using the OOD Audio Expression Generator, we can generate facial expressions by generating Blendshape coefficients through EmoTalk. As shown in Fig. \ref{fig:bs}, by using different Blendshapes, we can enable the character to display different expressions. Our method can effectively generate facial expressions continuously, consistently maintaining the character's identity without discontinuing issues between frames.

\begin{figure}
\begin{center}
   \includegraphics[width=1\linewidth]{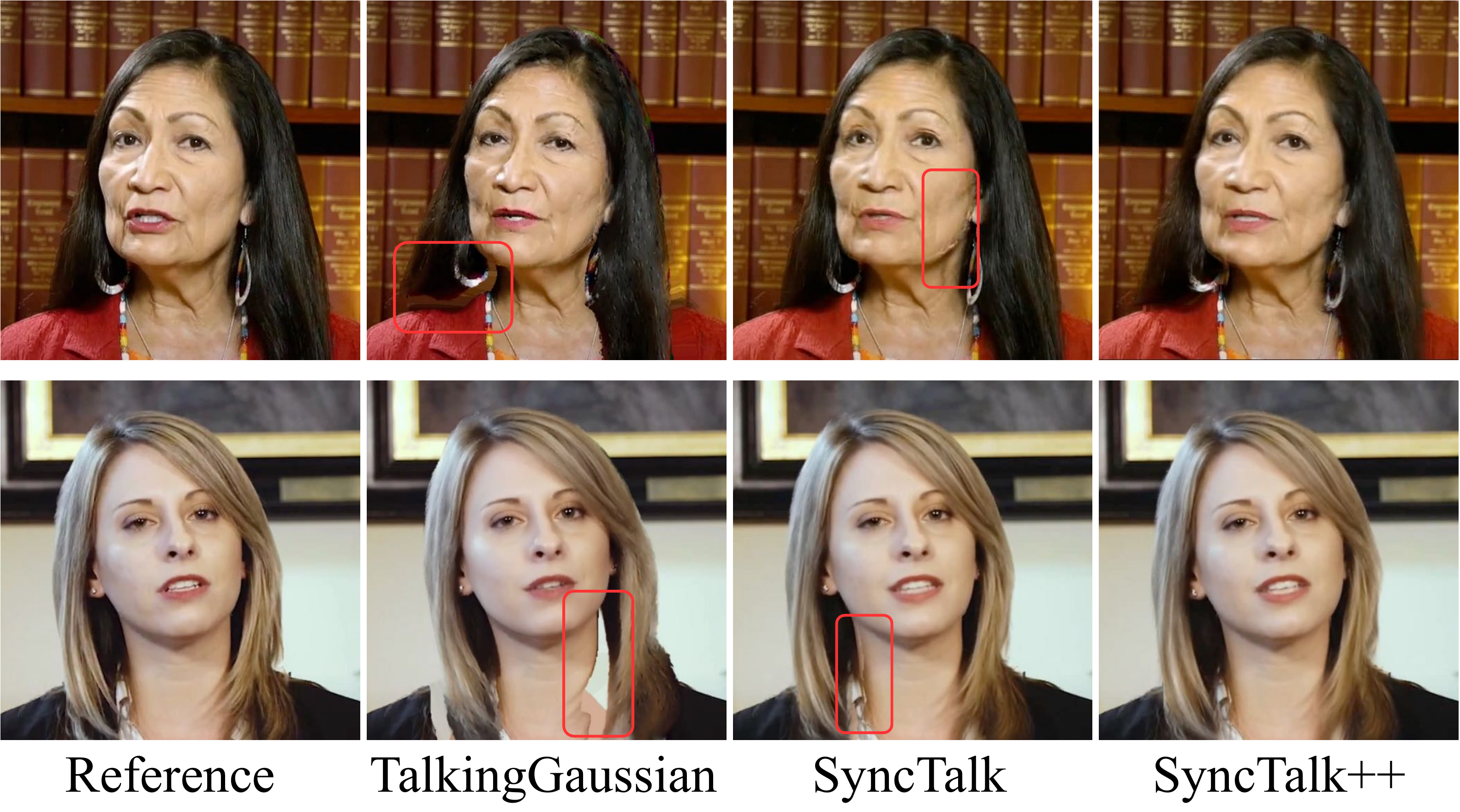}
\end{center}
   \caption{\textbf{Comparison of different trackers.} SyncTalk and TalkingGaussian trackers will cause obvious facial jitter and artifacts for long-haired characters, but SyncTalk++ will improve significantly.}
\label{fig:tracker}
\end{figure}

By incorporating the Semantic Weighting module, we have a more stable head tracker that enhances the stability of head poses. This improvement resulted in higher-quality reconstructions during training and significantly enhanced the visual coherence and realism of the generated videos. We compared our results with TalkingGaussian~\cite{li2024talkinggaussian} and SyncTalk~\cite{peng2024synctalk}, finding that our more stable tracker exhibited better visual quality, as shown in Fig.~\ref{fig:tracker}.

\noindent\textbf{User Study.} To assess the perceptual quality of our method, we conduct a comprehensive user study comparing SyncTalk++ with state-of-the-art approaches. We curate a dataset of 65 video clips, each lasting over 10 seconds, encompassing various head poses, facial expressions, and lip movements. Each method is represented by five clips. A total of 42 participants evaluate the videos, with an average completion time of 24 minutes per questionnaire. The study achieves a high reliability score, with a standardized Cronbach’s $\alpha$ coefficient of 0.96, ensuring the consistency of responses. The questionnaire follows the Mean Opinion Score (MOS) protocol, where participants rate the generated videos across five key aspects: (1) Lip-sync Accuracy, (2) Expression-sync Accuracy, (3) Pose-sync Accuracy, (4) Image Quality, and (5) Video Realness.

As shown in Table~\ref{tab:User}, SyncTalk++ consistently achieves the highest scores across all five metrics. Specifically, SyncTalk++ attains a Lip-sync Accuracy score of 4.309, outperforming the second-best SyncTalk by a 0.178 margin. For Expression-sync Accuracy, our method scores 4.154, exceeding IP-LAP and GaussianTalker. Additionally, SyncTalk++ achieves the best Pose-sync Accuracy at 4.371, a notable 0.268 improvement over IP-LAP.
In terms of visual quality, SyncTalk++ achieves an Image Quality score of 4.297, surpassing the second-best SyncTalk. Furthermore, it leads in Video Realness, scoring 4.229, which is 7.7$\%$ higher than TalkingGaussian.
In general, our approach significantly improves lip synchronization, expression synchronization, and pose alignment, while also improving image fidelity and video realism.

\subsection{Ablation Study}
We conduct an ablation study to systematically evaluate the contributions of different components in our model to the overall performance. To this end, we selected three core metrics for evaluation: Peak Signal-to-Noise Ratio (PSNR), Learned Perceptual Image Patch Similarity (LPIPS), and Landmark Distance (LMD). These metrics respectively measure image reconstruction quality, perceptual consistency, and the accuracy of lip synchronization. For testing, we chose a subject named “May,” and the results are presented in Table~\ref{tab:4}.

First, the Audio-Visual Encoder plays a critical role in the model, providing the primary lip sync information. When this module is replaced, we observe a significant deterioration in all three metrics, particularly with a 19.7$\%$ increase in the LMD error. This increase clearly indicates a decline in lip motion synchronization, further validating the importance of our Audio-Visual Encoder in extracting accurate audio features. This result underscores the ability of the Audio-Visual Encoder to capture fine lip movements synchronized with speech, which is crucial for generating realistic talking heads.

Next, we examine the impact of the Facial Animation Capture module, which captures facial expressions by using facial features. When this module is replaced with the AU units blink module, the metrics also worsen: PSNR decreases to 37.264, LPIPS rises to 0.0249, and LMD increases to 3.058. This suggests that the Facial Animation Capture module not only plays a vital role in lip synchronization but is also crucial for maintaining the naturalness and coherence of facial expressions.

The ablation of the Head-Sync Stabilizer further reveals its key role in reducing head pose jitter and preventing the separation of the head from the torso. Without this module, all metrics significantly decline: PSNR decreases to 29.193, LPIPS increases to 0.0749, and LMD rises to 3.264. This phenomenon indicates that the Head-Sync Stabilizer is essential for ensuring the stability of head movements and the overall consistency of the image.

The Portrait-Sync Generator focuses on restoring facial details. When this module is removed, noticeable segmentation boundaries appear in the generated images, particularly in the hair region. The ablation of the Semantic Weighting module reveals its importance in enhancing video stability. Removing this module results in a decline in all metrics, indicating its contribution to maintaining head pose stability in dynamic scenes.

\begin{figure}
\begin{center}
   \includegraphics[width=1\linewidth]{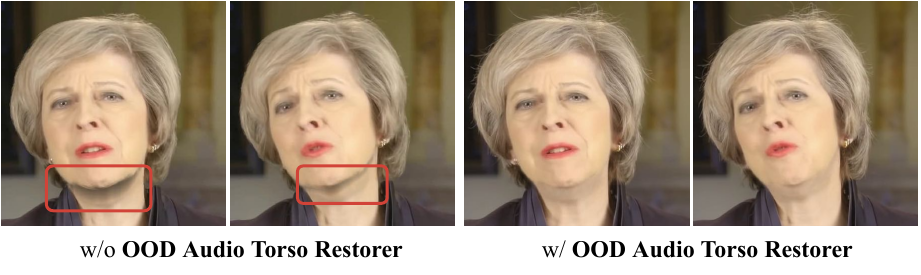}
\end{center}
   \caption{\textbf{Ablation study of OOD Audio Torso Restorer} Without using the Torso Restorer, there will be obvious pixel missing problems, and our method can repair them well.}
\label{fig:inpaint_compare}
\end{figure}

In addition, we conduct a dedicated ablation study on the OOD Audio Torso Restorer. When using OOD audio inputs during inference, the Torso Restorer effectively closes any pixel gaps between the generated head and the original torso, eliminating unnatural seams in the video. As shown in Tab.\ref{tab:Torso Restorer}, we evaluated three no-reference image quality metrics and observed a significant improvement after applying the Torso Restorer. Furthermore, Fig.\ref{fig:inpaint_compare} demonstrates that using the Torso Restorer markedly enhances visual quality and maintains coherence in the transition area between the face and torso.

\begin{table}
\setlength\tabcolsep{3pt}
\renewcommand{\arraystretch}{1.1}

\begin{center}
\resizebox{\linewidth}{!}{
\begin{tabular}{@{}llll@{}}
\toprule
                                                                                          & PSNR ↑ & LPIPS ↓ & LMD ↓  \\ \midrule
full                                                                                     & \textbf{39.093} & \textbf{0.0110}  & \textbf{2.715} \\ \midrule
\begin{tabular}[c]{@{}l@{}}replace Audio-Visual Encoder\\ with Hubert~\cite{hsu2021hubert}\end{tabular}     & 37.638 & 0.0194  & 3.250 \\
\begin{tabular}[c]{@{}l@{}}replace Facial Animation Capture\\ with AU units\end{tabular} & 37.264 & 0.0249  & 3.058 \\
w/o Head-Sync Stabilizer                                                                  & 29.193 & 0.0749  & 3.264 \\
w/o Portrait-Sync Generator                                                               & 33.759 & 0.0321  & 2.772 \\
w/o Semantic Weighting                                                                  & 34.526 & 0.0252  & 3.104 \\
\bottomrule
\end{tabular}
}
\end{center}
\caption{\textbf{Ablation study for our components.} We show the PSNR, LPIPS, and LMD in different cases.}
\label{tab:4}
\end{table}

\begin{table}[]
\begin{center}
\resizebox{\linewidth}{!}{
\begin{tabular}{@{}lcccc@{}}
\toprule
                           & NIQE ↓           & BRISQUE ↓         & HyperIQA ↑          \\ \midrule
SyncTalk++  (w/o Torso Restorer) & 15.2476          & 33.2916          & 62.0307                    \\
SyncTalk++ (w/ Torso Restorer)    & \textbf{14.3012} & \textbf{25.4147} & \textbf{66.1958}  \\ \bottomrule
\end{tabular}}
\end{center}
\caption{\textbf{Quantitative Results of the Torso Restorer.} Our Torso Restorer significantly improves image quality at OOD Audio settings. }
\label{tab:Torso Restorer}
\end{table}

\section{Ethical Consideration}
\label{sec:ethical}
Our SyncTalk and SyncTalk++ can synthesize high-quality, high-fidelity, audio-motion synchronized, visually indistinguishable talking-head videos. They are expected to contribute to developing fields such as human-computer interaction, artistic creation, digital agents, and digital twins. However, we must be aware that this type of deepfake talking-head video synthesis technology can be exploited for harmful purposes. In light of this, we have put forward a series of suggestions to try to mitigate the abuse of deepfake technology.

\noindent\textbf{Improve deepfake detection algorithms.} In recent years, there has been considerable work on detecting tampered videos, such as face swapping and reenactment \cite{dong2022protecting,guarnera2020deepfake,tolosana2020deepfakes}. However, distinguishing high-quality synthetic portraits based on recent NeRF and Gaussian Splatting methods remains challenging. We will share our work with the deepfake detection community, hoping it can help them develop more robust algorithms. Additionally, we attempt to distinguish the authenticity of videos based on rendering defects of NeRF and Gaussian Splatting. For example,
Gaussian Splatting-rendered novel-angle talking heads may show some unreasonable pixel points due to the incomplete convergence of 3D Gaussians.

\noindent\textbf{Protect real talking-head videos.} Since current methods based on NeRF and Gaussian Splatting strongly rely on real training videos, protecting them helps reduce the misuse of technology. For example, video and social media sites should take measures to prevent unauthorized video downloads or add digital watermarks to the portrait parts to interfere with training.

\noindent\textbf{Transparency and Consent.} In scenarios involving generating synthetic images or videos of individuals, explicit consent must be obtained. This includes informing participants about the nature of the technology, its capabilities, and the specific ways in which their likeness will be used. Transparency in the use of synthetic media is not just a legal obligation but a moral imperative to maintain trust and integrity in digital content.

\noindent\textbf{Restrict the application of deepfake technology.} The public should be made aware of the potential dangers of deepfake technology and urged to treat it cautiously. Additionally, we suggest establishing relevant laws to regulate the use of deepfake technology.

\section{Conclusion}
\label{sec:conclusion}
In this paper, we presented SyncTalk++, a framework for generating realistic speech-driven talking heads. By utilizing Gaussian Splatting, we achieved a rendering speed of 101 FPS and significantly reduced artifacts. The integration of the Expression Generator can generate speech-matched facial expressions, while the Torso Restorer addresses facial inconsistencies. Our experiments and user studies demonstrate that SyncTalk++ outperforms previous methods in synchronization, visual quality, and efficiency. These advancements pave the way for more immersive applications in digital assistants, virtual reality, and filmmaking.





\bibliographystyle{IEEEtran}
\bibliography{egbib}

\begin{thebibliography}{10}
\providecommand{\url}[1]{#1}
\csname url@samestyle\endcsname
\providecommand{\newblock}{\relax}
\providecommand{\bibinfo}[2]{#2}
\providecommand{\BIBentrySTDinterwordspacing}{\spaceskip=0pt\relax}
\providecommand{\BIBentryALTinterwordstretchfactor}{4}
\providecommand{\BIBentryALTinterwordspacing}{\spaceskip=\fontdimen2\font plus
\BIBentryALTinterwordstretchfactor\fontdimen3\font minus \fontdimen4\font\relax}
\providecommand{\BIBforeignlanguage}[2]{{%
\expandafter\ifx\csname l@#1\endcsname\relax
\typeout{** WARNING: IEEEtran.bst: No hyphenation pattern has been}%
\typeout{** loaded for the language `#1'. Using the pattern for}%
\typeout{** the default language instead.}%
\else
\language=\csname l@#1\endcsname
\fi
#2}}
\providecommand{\BIBdecl}{\relax}
\BIBdecl

\bibitem{thies2020neural}
J.~Thies, M.~Elgharib, A.~Tewari, C.~Theobalt, and M.~Nie{\ss}ner, ``Neural voice puppetry: Audio-driven facial reenactment,'' in \emph{Computer Vision--ECCV 2020: 16th European Conference, Glasgow, UK, August 23--28, 2020, Proceedings, Part XVI 16}.\hskip 1em plus 0.5em minus 0.4em\relax Springer, 2020, pp. 716--731.

\bibitem{yang2025megadance}
K.~Yang, X.~Tang, Z.~Peng, Y.~Hu, J.~He, and H.~Liu, ``Megadance: Mixture-of-experts architecture for genre-aware 3d dance generation,'' \emph{arXiv preprint arXiv:2505.17543}, 2025.

\bibitem{peng2023selftalk}
Z.~Peng, Y.~Luo, Y.~Shi, H.~Xu, X.~Zhu, H.~Liu, J.~He, and Z.~Fan, ``Selftalk: A self-supervised commutative training diagram to comprehend 3d talking faces,'' in \emph{Proceedings of the 31st ACM International Conference on Multimedia}, 2023, pp. 5292--5301.

\bibitem{zhou2024meta}
X.~Zhou, F.~Li, Z.~Peng, K.~Wu, J.~He, B.~Qin, Z.~Fan, and H.~Liu, ``Meta-learning empowered meta-face: Personalized speaking style adaptation for audio-driven 3d talking face animation,'' \emph{arXiv preprint arXiv:2408.09357}, 2024.

\bibitem{kim2018deep}
H.~Kim, P.~Garrido, A.~Tewari, W.~Xu, J.~Thies, M.~Niessner, P.~P{\'e}rez, C.~Richardt, M.~Zollh{\"o}fer, and C.~Theobalt, ``Deep video portraits,'' \emph{ACM transactions on graphics (TOG)}, vol.~37, no.~4, pp. 1--14, 2018.

\bibitem{peng2025dualtalk}
Z.~Peng, Y.~Fan, H.~Wu, X.~Wang, H.~Liu, J.~He, and Z.~Fan, ``Dualtalk: Dual-speaker interaction for 3d talking head conversations,'' in \emph{Proceedings of the Computer Vision and Pattern Recognition Conference}, 2025, pp. 21\,055--21\,064.

\bibitem{wu2024vgg}
H.~Wu, Z.~Peng, X.~Zhou, Y.~Cheng, J.~He, H.~Liu, and Z.~Fan, ``Vgg-tex: A vivid geometry-guided facial texture estimation model for high fidelity monocular 3d face reconstruction,'' \emph{arXiv preprint arXiv:2409.09740}, 2024.

\bibitem{goodfellow2020generative}
I.~Goodfellow, J.~Pouget-Abadie, M.~Mirza, B.~Xu, D.~Warde-Farley, S.~Ozair, A.~Courville, and Y.~Bengio, ``Generative adversarial networks,'' \emph{Communications of the ACM}, vol.~63, no.~11, pp. 139--144, 2020.

\bibitem{zhang2023dinet}
Z.~Zhang, Z.~Hu, W.~Deng, C.~Fan, T.~Lv, and Y.~Ding, ``Dinet: Deformation inpainting network for realistic face visually dubbing on high resolution video,'' \emph{arXiv preprint arXiv:2303.03988}, 2023.

\bibitem{wang2023seeing}
J.~Wang, X.~Qian, M.~Zhang, R.~T. Tan, and H.~Li, ``Seeing what you said: Talking face generation guided by a lip reading expert,'' in \emph{Proceedings of the IEEE/CVF Conference on Computer Vision and Pattern Recognition}, 2023, pp. 14\,653--14\,662.

\bibitem{guan2023stylesync}
J.~Guan, Z.~Zhang, H.~Zhou, T.~Hu, K.~Wang, D.~He, H.~Feng, J.~Liu, E.~Ding, Z.~Liu \emph{et~al.}, ``Stylesync: High-fidelity generalized and personalized lip sync in style-based generator,'' in \emph{Proceedings of the IEEE/CVF Conference on Computer Vision and Pattern Recognition}, 2023, pp. 1505--1515.

\bibitem{zhong2023identity}
W.~Zhong, C.~Fang, Y.~Cai, P.~Wei, G.~Zhao, L.~Lin, and G.~Li, ``Identity-preserving talking face generation with landmark and appearance priors,'' in \emph{Proceedings of the IEEE/CVF Conference on Computer Vision and Pattern Recognition}, 2023, pp. 9729--9738.

\bibitem{tan2023emmn}
S.~Tan, B.~Ji, and Y.~Pan, ``Emmn: Emotional motion memory network for audio-driven emotional talking face generation,'' in \emph{Proceedings of the IEEE/CVF International Conference on Computer Vision}, 2023, pp. 22\,146--22\,156.

\bibitem{guo2024liveportrait}
J.~Guo, D.~Zhang, X.~Liu, Z.~Zhong, Y.~Zhang, P.~Wan, and D.~Zhang, ``Liveportrait: Efficient portrait animation with stitching and retargeting control,'' \emph{arXiv preprint arXiv:2407.03168}, 2024.

\bibitem{tan2024flowvqtalker}
S.~Tan, B.~Ji, and Y.~Pan, ``Flowvqtalker: High-quality emotional talking face generation through normalizing flow and quantization,'' in \emph{Proceedings of the IEEE/CVF Conference on Computer Vision and Pattern Recognition}, 2024, pp. 26\,317--26\,327.

\bibitem{wu2023ganhead}
S.~Wu, Y.~Yan, Y.~Li, Y.~Cheng, W.~Zhu, K.~Gao, X.~Li, and G.~Zhai, ``Ganhead: Towards generative animatable neural head avatars,'' in \emph{Proceedings of the IEEE/CVF conference on computer vision and pattern recognition}, 2023, pp. 437--447.

\bibitem{tian2024emo}
L.~Tian, Q.~Wang, B.~Zhang, and L.~Bo, ``Emo: Emote portrait alive-generating expressive portrait videos with audio2video diffusion model under weak conditions,'' \emph{arXiv preprint arXiv:2402.17485}, 2024.

\bibitem{ma2023dreamtalk}
Y.~Ma, S.~Zhang, J.~Wang, X.~Wang, Y.~Zhang, and Z.~Deng, ``Dreamtalk: When expressive talking head generation meets diffusion probabilistic models,'' \emph{arXiv preprint arXiv:2312.09767}, 2023.

\bibitem{shen2023difftalk}
S.~Shen, W.~Zhao, Z.~Meng, W.~Li, Z.~Zhu, J.~Zhou, and J.~Lu, ``Difftalk: Crafting diffusion models for generalized audio-driven portraits animation,'' in \emph{Proceedings of the IEEE/CVF Conference on Computer Vision and Pattern Recognition}, 2023, pp. 1982--1991.

\bibitem{mildenhall2021nerf}
B.~Mildenhall, P.~P. Srinivasan, M.~Tancik, J.~T. Barron, R.~Ramamoorthi, and R.~Ng, ``Nerf: Representing scenes as neural radiance fields for view synthesis,'' \emph{Communications of the ACM}, vol.~65, no.~1, pp. 99--106, 2021.

\bibitem{guo2021ad}
Y.~Guo, K.~Chen, S.~Liang, Y.-J. Liu, H.~Bao, and J.~Zhang, ``Ad-nerf: Audio driven neural radiance fields for talking head synthesis,'' in \emph{Proceedings of the IEEE/CVF International Conference on Computer Vision}, 2021, pp. 5784--5794.

\bibitem{yao2022dfa}
S.~Yao, R.~Zhong, Y.~Yan, G.~Zhai, and X.~Yang, ``Dfa-nerf: Personalized talking head generation via disentangled face attributes neural rendering,'' \emph{arXiv preprint arXiv:2201.00791}, 2022.

\bibitem{shen2022learning}
S.~Shen, W.~Li, Z.~Zhu, Y.~Duan, J.~Zhou, and J.~Lu, ``Learning dynamic facial radiance fields for few-shot talking head synthesis,'' in \emph{European Conference on Computer Vision}.\hskip 1em plus 0.5em minus 0.4em\relax Springer, 2022, pp. 666--682.

\bibitem{ye2023geneface}
Z.~Ye, Z.~Jiang, Y.~Ren, J.~Liu, J.~He, and Z.~Zhao, ``Geneface: Generalized and high-fidelity audio-driven 3d talking face synthesis,'' \emph{arXiv preprint arXiv:2301.13430}, 2023.

\bibitem{li2023efficient}
J.~Li, J.~Zhang, X.~Bai, J.~Zhou, and L.~Gu, ``Efficient region-aware neural radiance fields for high-fidelity talking portrait synthesis,'' in \emph{Proceedings of the IEEE/CVF International Conference on Computer Vision}, 2023, pp. 7568--7578.

\bibitem{zhang2024learning}
Z.~Zhang, R.~Zheng, B.~Li, C.~Han, T.~Li, M.~Wang, T.~Guo, J.~Chen, Z.~Liu, and M.~Yang, ``Learning dynamic tetrahedra for high-quality talking head synthesis,'' in \emph{Proceedings of the IEEE/CVF Conference on Computer Vision and Pattern Recognition}, 2024, pp. 5209--5219.

\bibitem{chu2024gpavatar}
X.~Chu, Y.~Li, A.~Zeng, T.~Yang, L.~Lin, Y.~Liu, and T.~Harada, ``Gpavatar: Generalizable and precise head avatar from image (s),'' \emph{arXiv preprint arXiv:2401.10215}, 2024.

\bibitem{li2024er}
J.~Li, J.~Zhang, X.~Bai, J.~Zheng, J.~Zhou, and L.~Gu, ``Er-nerf++: Efficient region-aware neural radiance fields for high-fidelity talking portrait synthesis,'' \emph{Information Fusion}, vol. 110, p. 102456, 2024.

\bibitem{kerbl20233d}
B.~Kerbl, G.~Kopanas, T.~Leimk{\"u}hler, and G.~Drettakis, ``3d gaussian splatting for real-time radiance field rendering.'' \emph{ACM Trans. Graph.}, vol.~42, no.~4, pp. 139--1, 2023.

\bibitem{yu2024gaussiantalker}
H.~Yu, Z.~Qu, Q.~Yu, J.~Chen, Z.~Jiang, Z.~Chen, S.~Zhang, J.~Xu, F.~Wu, C.~Lv \emph{et~al.}, ``Gaussiantalker: Speaker-specific talking head synthesis via 3d gaussian splatting,'' \emph{arXiv preprint arXiv:2404.14037}, 2024.

\bibitem{li2024talkinggaussian}
J.~Li, J.~Zhang, X.~Bai, J.~Zheng, X.~Ning, J.~Zhou, and L.~Gu, ``Talkinggaussian: Structure-persistent 3d talking head synthesis via gaussian splatting,'' \emph{arXiv preprint arXiv:2404.15264}, 2024.

\bibitem{amodei2016deep}
D.~Amodei, S.~Ananthanarayanan, R.~Anubhai, J.~Bai, E.~Battenberg, C.~Case, J.~Casper, B.~Catanzaro, Q.~Cheng, G.~Chen \emph{et~al.}, ``Deep speech 2: End-to-end speech recognition in english and mandarin,'' in \emph{International conference on machine learning}.\hskip 1em plus 0.5em minus 0.4em\relax PMLR, 2016, pp. 173--182.

\bibitem{peng2024synctalk}
Z.~Peng, W.~Hu, Y.~Shi, X.~Zhu, X.~Zhang, H.~Zhao, J.~He, H.~Liu, and Z.~Fan, ``Synctalk: The devil is in the synchronization for talking head synthesis,'' in \emph{Proceedings of the IEEE/CVF Conference on Computer Vision and Pattern Recognition}, 2024, pp. 666--676.

\bibitem{chen2018lip}
L.~Chen, Z.~Li, R.~K. Maddox, Z.~Duan, and C.~Xu, ``Lip movements generation at a glance,'' in \emph{Proceedings of the European conference on computer vision (ECCV)}, 2018, pp. 520--535.

\bibitem{kr2019towards}
P.~KR, R.~Mukhopadhyay, J.~Philip, A.~Jha, V.~Namboodiri, and C.~Jawahar, ``Towards automatic face-to-face translation,'' in \emph{Proceedings of the 27th ACM international conference on multimedia}, 2019, pp. 1428--1436.

\bibitem{chen2019hierarchical}
L.~Chen, R.~K. Maddox, Z.~Duan, and C.~Xu, ``Hierarchical cross-modal talking face generation with dynamic pixel-wise loss,'' in \emph{Proceedings of the IEEE/CVF conference on computer vision and pattern recognition}, 2019, pp. 7832--7841.

\bibitem{zhou2019talking}
H.~Zhou, Y.~Liu, Z.~Liu, P.~Luo, and X.~Wang, ``Talking face generation by adversarially disentangled audio-visual representation,'' in \emph{Proceedings of the AAAI conference on artificial intelligence}, vol.~33, no.~01, 2019, pp. 9299--9306.

\bibitem{das2020speech}
D.~Das, S.~Biswas, S.~Sinha, and B.~Bhowmick, ``Speech-driven facial animation using cascaded gans for learning of motion and texture,'' in \emph{Computer Vision--ECCV 2020: 16th European Conference, Glasgow, UK, August 23--28, 2020, Proceedings, Part XXX 16}.\hskip 1em plus 0.5em minus 0.4em\relax Springer, 2020, pp. 408--424.

\bibitem{vougioukas2020realistic}
K.~Vougioukas, S.~Petridis, and M.~Pantic, ``Realistic speech-driven facial animation with gans,'' \emph{International Journal of Computer Vision}, vol. 128, pp. 1398--1413, 2020.

\bibitem{meshry2021learned}
M.~Meshry, S.~Suri, L.~S. Davis, and A.~Shrivastava, ``Learned spatial representations for few-shot talking-head synthesis,'' in \emph{Proceedings of the IEEE/CVF International Conference on Computer Vision}, 2021, pp. 13\,829--13\,838.

\bibitem{zhou2021pose}
H.~Zhou, Y.~Sun, W.~Wu, C.~C. Loy, X.~Wang, and Z.~Liu, ``Pose-controllable talking face generation by implicitly modularized audio-visual representation,'' in \emph{Proceedings of the IEEE/CVF conference on computer vision and pattern recognition}, 2021, pp. 4176--4186.

\bibitem{song2022everybody}
L.~Song, W.~Wu, C.~Qian, R.~He, and C.~C. Loy, ``Everybody’s talkin’: Let me talk as you want,'' \emph{IEEE Transactions on Information Forensics and Security}, vol.~17, pp. 585--598, 2022.

\bibitem{prajwal2020lip}
K.~Prajwal, R.~Mukhopadhyay, V.~P. Namboodiri, and C.~Jawahar, ``A lip sync expert is all you need for speech to lip generation in the wild,'' in \emph{Proceedings of the 28th ACM international conference on multimedia}, 2020, pp. 484--492.

\bibitem{zhou2020makelttalk}
Y.~Zhou, X.~Han, E.~Shechtman, J.~Echevarria, E.~Kalogerakis, and D.~Li, ``Makelttalk: speaker-aware talking-head animation,'' \emph{ACM Transactions On Graphics (TOG)}, vol.~39, no.~6, pp. 1--15, 2020.

\bibitem{lu2021live}
Y.~Lu, J.~Chai, and X.~Cao, ``Live speech portraits: real-time photorealistic talking-head animation,'' \emph{ACM Transactions on Graphics (TOG)}, vol.~40, no.~6, pp. 1--17, 2021.

\bibitem{wang2021audio2head}
S.~Wang, L.~Li, Y.~Ding, C.~Fan, and X.~Yu, ``Audio2head: Audio-driven one-shot talking-head generation with natural head motion,'' \emph{arXiv preprint arXiv:2107.09293}, 2021.

\bibitem{zhang2023sadtalker}
W.~Zhang, X.~Cun, X.~Wang, Y.~Zhang, X.~Shen, Y.~Guo, Y.~Shan, and F.~Wang, ``Sadtalker: Learning realistic 3d motion coefficients for stylized audio-driven single image talking face animation,'' in \emph{Proceedings of the IEEE/CVF Conference on Computer Vision and Pattern Recognition}, 2023, pp. 8652--8661.

\bibitem{xu2024hallo}
M.~Xu, H.~Li, Q.~Su, H.~Shang, L.~Zhang, C.~Liu, J.~Wang, L.~Van~Gool, Y.~Yao, and S.~Zhu, ``Hallo: Hierarchical audio-driven visual synthesis for portrait image animation,'' \emph{arXiv preprint arXiv:2406.08801}, 2024.

\bibitem{chen2024echomimic}
Z.~Chen, J.~Cao, Z.~Chen, Y.~Li, and C.~Ma, ``Echomimic: Lifelike audio-driven portrait animations through editable landmark conditions,'' \emph{arXiv preprint arXiv:2407.08136}, 2024.

\bibitem{peng2025omnisync}
Z.~Peng, J.~Liu, H.~Zhang, X.~Liu, S.~Tang, P.~Wan, D.~Zhang, H.~Liu, and J.~He, ``Omnisync: Towards universal lip synchronization via diffusion transformers,'' \emph{arXiv preprint arXiv:2505.21448}, 2025.

\bibitem{rombach2022high}
R.~Rombach, A.~Blattmann, D.~Lorenz, P.~Esser, and B.~Ommer, ``High-resolution image synthesis with latent diffusion models,'' in \emph{Proceedings of the IEEE/CVF conference on computer vision and pattern recognition}, 2022, pp. 10\,684--10\,695.

\bibitem{martin2021nerf}
R.~Martin-Brualla, N.~Radwan, M.~S. Sajjadi, J.~T. Barron, A.~Dosovitskiy, and D.~Duckworth, ``Nerf in the wild: Neural radiance fields for unconstrained photo collections,'' in \emph{Proceedings of the IEEE/CVF Conference on Computer Vision and Pattern Recognition}, 2021, pp. 7210--7219.

\bibitem{gao2022nerf}
K.~Gao, Y.~Gao, H.~He, D.~Lu, L.~Xu, and J.~Li, ``Nerf: Neural radiance field in 3d vision, a comprehensive review,'' \emph{arXiv preprint arXiv:2210.00379}, 2022.

\bibitem{liu2022semantic}
X.~Liu, Y.~Xu, Q.~Wu, H.~Zhou, W.~Wu, and B.~Zhou, ``Semantic-aware implicit neural audio-driven video portrait generation,'' in \emph{European Conference on Computer Vision}.\hskip 1em plus 0.5em minus 0.4em\relax Springer, 2022, pp. 106--125.

\bibitem{tang2022real}
J.~Tang, K.~Wang, H.~Zhou, X.~Chen, D.~He, T.~Hu, J.~Liu, G.~Zeng, and J.~Wang, ``Real-time neural radiance talking portrait synthesis via audio-spatial decomposition,'' \emph{arXiv preprint arXiv:2211.12368}, 2022.

\bibitem{muller2022instant}
T.~M{\"u}ller, A.~Evans, C.~Schied, and A.~Keller, ``Instant neural graphics primitives with a multiresolution hash encoding,'' \emph{ACM transactions on graphics (TOG)}, vol.~41, no.~4, pp. 1--15, 2022.

\bibitem{deng2024portrait4d}
Y.~Deng, D.~Wang, X.~Ren, X.~Chen, and B.~Wang, ``Portrait4d: Learning one-shot 4d head avatar synthesis using synthetic data,'' in \emph{Proceedings of the IEEE/CVF Conference on Computer Vision and Pattern Recognition}, 2024, pp. 7119--7130.

\bibitem{gafni2021dynamic}
G.~Gafni, J.~Thies, M.~Zollhofer, and M.~Nie{\ss}ner, ``Dynamic neural radiance fields for monocular 4d facial avatar reconstruction,'' in \emph{Proceedings of the IEEE/CVF Conference on Computer Vision and Pattern Recognition}, 2021, pp. 8649--8658.

\bibitem{zheng2022avatar}
Y.~Zheng, V.~F. Abrevaya, M.~C. B{\"u}hler, X.~Chen, M.~J. Black, and O.~Hilliges, ``Im avatar: Implicit morphable head avatars from videos,'' in \emph{Proceedings of the IEEE/CVF Conference on Computer Vision and Pattern Recognition}, 2022, pp. 13\,545--13\,555.

\bibitem{zielonka2023instant}
W.~Zielonka, T.~Bolkart, and J.~Thies, ``Instant volumetric head avatars,'' in \emph{Proceedings of the IEEE/CVF Conference on Computer Vision and Pattern Recognition}, 2023, pp. 4574--4584.

\bibitem{zheng2023pointavatar}
Y.~Zheng, W.~Yifan, G.~Wetzstein, M.~J. Black, and O.~Hilliges, ``Pointavatar: Deformable point-based head avatars from videos,'' in \emph{Proceedings of the IEEE/CVF Conference on Computer Vision and Pattern Recognition}, 2023, pp. 21\,057--21\,067.

\bibitem{lu2024scaffold}
T.~Lu, M.~Yu, L.~Xu, Y.~Xiangli, L.~Wang, D.~Lin, and B.~Dai, ``Scaffold-gs: Structured 3d gaussians for view-adaptive rendering,'' in \emph{Proceedings of the IEEE/CVF Conference on Computer Vision and Pattern Recognition}, 2024, pp. 20\,654--20\,664.

\bibitem{yu2024mip}
Z.~Yu, A.~Chen, B.~Huang, T.~Sattler, and A.~Geiger, ``Mip-splatting: Alias-free 3d gaussian splatting,'' in \emph{Proceedings of the IEEE/CVF Conference on Computer Vision and Pattern Recognition}, 2024, pp. 19\,447--19\,456.

\bibitem{qian20243dgs}
Z.~Qian, S.~Wang, M.~Mihajlovic, A.~Geiger, and S.~Tang, ``3dgs-avatar: Animatable avatars via deformable 3d gaussian splatting,'' in \emph{Proceedings of the IEEE/CVF Conference on Computer Vision and Pattern Recognition}, 2024, pp. 5020--5030.

\bibitem{hu2024gauhuman}
S.~Hu, T.~Hu, and Z.~Liu, ``Gauhuman: Articulated gaussian splatting from monocular human videos,'' in \emph{Proceedings of the IEEE/CVF Conference on Computer Vision and Pattern Recognition}, 2024, pp. 20\,418--20\,431.

\bibitem{zhao2024psavatar}
Z.~Zhao, Z.~Bao, Q.~Li, G.~Qiu, and K.~Liu, ``Psavatar: A point-based morphable shape model for real-time head avatar creation with 3d gaussian splatting,'' \emph{arXiv preprint arXiv:2401.12900}, 2024.

\bibitem{qian2024gaussianavatars}
S.~Qian, T.~Kirschstein, L.~Schoneveld, D.~Davoli, S.~Giebenhain, and M.~Nie{\ss}ner, ``Gaussianavatars: Photorealistic head avatars with rigged 3d gaussians,'' in \emph{Proceedings of the IEEE/CVF Conference on Computer Vision and Pattern Recognition}, 2024, pp. 20\,299--20\,309.

\bibitem{xu2024gaussian}
Y.~Xu, B.~Chen, Z.~Li, H.~Zhang, L.~Wang, Z.~Zheng, and Y.~Liu, ``Gaussian head avatar: Ultra high-fidelity head avatar via dynamic gaussians,'' in \emph{Proceedings of the IEEE/CVF Conference on Computer Vision and Pattern Recognition}, 2024, pp. 1931--1941.

\bibitem{cho2024gaussiantalker}
K.~Cho, J.~Lee, H.~Yoon, Y.~Hong, J.~Ko, S.~Ahn, and S.~Kim, ``Gaussiantalker: Real-time high-fidelity talking head synthesis with audio-driven 3d gaussian splatting,'' \emph{arXiv preprint arXiv:2404.16012}, 2024.

\bibitem{baevski2020wav2vec}
A.~Baevski, Y.~Zhou, A.~Mohamed, and M.~Auli, ``wav2vec 2.0: A framework for self-supervised learning of speech representations,'' \emph{Advances in neural information processing systems}, vol.~33, pp. 12\,449--12\,460, 2020.

\bibitem{hsu2021hubert}
W.-N. Hsu, B.~Bolte, Y.-H.~H. Tsai, K.~Lakhotia, R.~Salakhutdinov, and A.~Mohamed, ``Hubert: Self-supervised speech representation learning by masked prediction of hidden units,'' \emph{IEEE/ACM Transactions on Audio, Speech, and Language Processing}, vol.~29, pp. 3451--3460, 2021.

\bibitem{afouras2018deep}
T.~Afouras, J.~S. Chung, A.~Senior, O.~Vinyals, and A.~Zisserman, ``Deep audio-visual speech recognition,'' \emph{IEEE transactions on pattern analysis and machine intelligence}, vol.~44, no.~12, pp. 8717--8727, 2018.

\bibitem{chung2017out}
J.~S. Chung and A.~Zisserman, ``Out of time: automated lip sync in the wild,'' in \emph{Computer Vision--ACCV 2016 Workshops: ACCV 2016 International Workshops, Taipei, Taiwan, November 20-24, 2016, Revised Selected Papers, Part II 13}.\hskip 1em plus 0.5em minus 0.4em\relax Springer, 2017, pp. 251--263.

\bibitem{peng2023emotalk}
Z.~Peng, H.~Wu, Z.~Song, H.~Xu, X.~Zhu, J.~He, H.~Liu, and Z.~Fan, ``Emotalk: Speech-driven emotional disentanglement for 3d face animation,'' in \emph{Proceedings of the IEEE/CVF International Conference on Computer Vision}, 2023, pp. 20\,687--20\,697.

\bibitem{he2016deep}
K.~He, X.~Zhang, S.~Ren, and J.~Sun, ``Deep residual learning for image recognition,'' in \emph{Proceedings of the IEEE conference on computer vision and pattern recognition}, 2016, pp. 770--778.

\bibitem{face-alignment}
A.~Bulat, ``face-alignment: 2d and 3d face alignment library build using pytorch,'' \url{https://github.com/1adrianb/face-alignment}, 2017.

\bibitem{paysan20093d}
P.~Paysan, R.~Knothe, B.~Amberg, S.~Romdhani, and T.~Vetter, ``A 3d face model for pose and illumination invariant face recognition,'' in \emph{2009 sixth IEEE international conference on advanced video and signal based surveillance}.\hskip 1em plus 0.5em minus 0.4em\relax Ieee, 2009, pp. 296--301.

\bibitem{kingma2014adam}
D.~P. Kingma and J.~Ba, ``Adam: A method for stochastic optimization,'' \emph{arXiv preprint arXiv:1412.6980}, 2014.

\bibitem{zwicker2001ewa}
M.~Zwicker, H.~Pfister, J.~Van~Baar, and M.~Gross, ``Ewa volume splatting,'' in \emph{Proceedings Visualization, 2001. VIS'01.}\hskip 1em plus 0.5em minus 0.4em\relax IEEE, 2001, pp. 29--538.

\bibitem{chan2022efficient}
E.~R. Chan, C.~Z. Lin, M.~A. Chan, K.~Nagano, B.~Pan, S.~De~Mello, O.~Gallo, L.~J. Guibas, J.~Tremblay, S.~Khamis \emph{et~al.}, ``Efficient geometry-aware 3d generative adversarial networks,'' in \emph{Proceedings of the IEEE/CVF conference on computer vision and pattern recognition}, 2022, pp. 16\,123--16\,133.

\bibitem{fridovich2023k}
S.~Fridovich-Keil, G.~Meanti, F.~R. Warburg, B.~Recht, and A.~Kanazawa, ``K-planes: Explicit radiance fields in space, time, and appearance,'' in \emph{Proceedings of the IEEE/CVF Conference on Computer Vision and Pattern Recognition}, 2023, pp. 12\,479--12\,488.

\bibitem{hu2023tri}
W.~Hu, Y.~Wang, L.~Ma, B.~Yang, L.~Gao, X.~Liu, and Y.~Ma, ``Tri-miprf: Tri-mip representation for efficient anti-aliasing neural radiance fields,'' in \emph{Proceedings of the IEEE/CVF International Conference on Computer Vision}, 2023, pp. 19\,774--19\,783.

\bibitem{van2017neural}
A.~Van Den~Oord, O.~Vinyals \emph{et~al.}, ``Neural discrete representation learning,'' \emph{Advances in neural information processing systems}, vol.~30, 2017.

\bibitem{cheng2022videoretalking}
K.~Cheng, X.~Cun, Y.~Zhang, M.~Xia, F.~Yin, M.~Zhu, X.~Wang, J.~Wang, and N.~Wang, ``Videoretalking: Audio-based lip synchronization for talking head video editing in the wild,'' in \emph{SIGGRAPH Asia 2022 Conference Papers}, 2022, pp. 1--9.

\bibitem{zhang2018unreasonable}
R.~Zhang, P.~Isola, A.~A. Efros, E.~Shechtman, and O.~Wang, ``The unreasonable effectiveness of deep features as a perceptual metric,'' in \emph{Proceedings of the IEEE conference on computer vision and pattern recognition}, 2018, pp. 586--595.

\bibitem{heusel2017gans}
M.~Heusel, H.~Ramsauer, T.~Unterthiner, B.~Nessler, and S.~Hochreiter, ``Gans trained by a two time-scale update rule converge to a local nash equilibrium,'' \emph{Advances in neural information processing systems}, vol.~30, 2017.

\bibitem{zhang2019ranksrgan}
W.~Zhang, Y.~Liu, C.~Dong, and Y.~Qiao, ``Ranksrgan: Generative adversarial networks with ranker for image super-resolution,'' in \emph{Proceedings of the IEEE/CVF International Conference on Computer Vision}, 2019, pp. 3096--3105.

\bibitem{mittal2012making}
A.~Mittal, R.~Soundararajan, and A.~C. Bovik, ``Making a “completely blind” image quality analyzer,'' \emph{IEEE Signal processing letters}, vol.~20, no.~3, pp. 209--212, 2012.

\bibitem{mittal2012no}
A.~Mittal, A.~K. Moorthy, and A.~C. Bovik, ``No-reference image quality assessment in the spatial domain,'' \emph{IEEE Transactions on image processing}, vol.~21, no.~12, pp. 4695--4708, 2012.

\bibitem{Su_2020_CVPR}
S.~Su, Q.~Yan, Y.~Zhu, C.~Zhang, X.~Ge, J.~Sun, and Y.~Zhang, ``Blindly assess image quality in the wild guided by a self-adaptive hyper network,'' in \emph{IEEE/CVF Conference on Computer Vision and Pattern Recognition (CVPR)}, June 2020.

\bibitem{baltruvsaitis2015cross}
T.~Baltru{\v{s}}aitis, M.~Mahmoud, and P.~Robinson, ``Cross-dataset learning and person-specific normalisation for automatic action unit detection,'' in \emph{2015 11th IEEE International Conference and Workshops on Automatic Face and Gesture Recognition (FG)}, vol.~6.\hskip 1em plus 0.5em minus 0.4em\relax IEEE, 2015, pp. 1--6.

\bibitem{dong2022protecting}
X.~Dong, J.~Bao, D.~Chen, T.~Zhang, W.~Zhang, N.~Yu, D.~Chen, F.~Wen, and B.~Guo, ``Protecting celebrities from deepfake with identity consistency transformer,'' in \emph{Proceedings of the IEEE/CVF Conference on Computer Vision and Pattern Recognition}, 2022, pp. 9468--9478.

\bibitem{guarnera2020deepfake}
L.~Guarnera, O.~Giudice, and S.~Battiato, ``Deepfake detection by analyzing convolutional traces,'' in \emph{Proceedings of the IEEE/CVF conference on computer vision and pattern recognition workshops}, 2020, pp. 666--667.

\bibitem{tolosana2020deepfakes}
R.~Tolosana, R.~Vera-Rodriguez, J.~Fierrez, A.~Morales, and J.~Ortega-Garcia, ``Deepfakes and beyond: A survey of face manipulation and fake detection,'' \emph{Information Fusion}, vol.~64, pp. 131--148, 2020.

\end{thebibliography}

\vfill

\end{document}